\documentclass{article} 

\usepackage{iclr2027_conference,times}
\iclrfinalcopy 

\usepackage{amsmath,amsfonts,bm}

\def\eqref#1{equation~\ref{#1}}

\def\1{\bm{1}}

\DeclareMathAlphabet{\mathsfit}{\encodingdefault}{\sfdefault}{m}{sl}
\SetMathAlphabet{\mathsfit}{bold}{\encodingdefault}{\sfdefault}{bx}{n}

\usepackage{hyperref}
\usepackage{url}
\usepackage{latexsym}
\usepackage[T1]{fontenc}
\usepackage[utf8]{inputenc}
\usepackage{microtype}
\usepackage{inconsolata}
\usepackage{graphicx}
\usepackage{booktabs}
\usepackage{makecell}
\usepackage{multirow}
\usepackage{amsmath}
\usepackage[frozencache,cachedir=_minted]{minted}
\usepackage{tcolorbox}
\usepackage{xcolor}
\usepackage{hyperref}
\usepackage{url}
\hypersetup{
  colorlinks=true,
  linkcolor=blue,
  citecolor=blue,
  urlcolor=blue
}
\usepackage{wrapfig}
\usepackage{caption}
\usepackage{enumitem}
\usepackage{float}

\newtcolorbox{promptbox}[1][]{
  colback=gray!4,
  colframe=gray!45,
  coltitle=black,
  arc=2pt,
  boxrule=0.4pt,
  left=5pt,
  right=5pt,
  top=4pt,
  bottom=4pt,
  fonttitle=\bfseries\scriptsize,
  fontupper=\scriptsize,
  title=#1
}

\title{EgoCITE: Context-Augmented Indexing and Time-Aware Retrieval for Long-Horizon \mbox{Egocentric} Memory}

\author{
Le Zhang\textsuperscript{1} \qquad
Hao Chen\textsuperscript{2} \qquad
Vlad Roznyatovskiy\textsuperscript{2} \qquad
Jianzhong Zhang\textsuperscript{2} \qquad
Ke Sun\textsuperscript{1} \\
\textsuperscript{1}University of Michigan, Ann Arbor \qquad
\textsuperscript{2}Samsung Research America \\
{\small\texttt{\{zle, kesuniot\}@umich.edu} \qquad
\texttt{\{hao.chen1, vlad.r, jianzhong.z\}@samsung.com}}
}

\begin{document}
\maketitle

\begin{abstract}

Long-horizon egocentric memory transforms continuous first-person video and audio into a searchable record of past experiences.
We demonstrate two bottlenecks in existing systems: indices built from context-poor captions are unreliable for agentic search, while retrieval ignores a question's temporal intent.
To address both bottlenecks, we introduce \texttt{EgoCITE} (Egocentric Context-augmented Indexing and Time-aware Evidence retrieval), a long-horizon agentic memory framework for egocentric QA.
\texttt{EgoCITE} comprises three components.
\texttt{EgoScheme} uses local multimodal context to turn fragmentary video captions and speech transcripts into self-contained atomic memory indices.
\texttt{EgoIndex} organizes complementary action, activity, utterance, and conversation representations into searchable multi-view memory indices at multiple granularities.
\texttt{EgoRetrv} combines semantic search with question-conditioned temporal relevance scoring and curation of retrieved evidence.
We evaluate \texttt{EgoCITE} on EgoLifeQA, EgoMem, and EgoR1-Bench in terms of answer accuracy and target-event retrieval alignment.
\texttt{EgoCITE} improves accuracy over agentic memory baselines by at least 4.4--14.2\% while achieving 36$\times$ lower cost than long-context LLM agents.

\end{abstract}


\section{Introduction}
\begin{wrapfigure}{r}{0.45\textwidth}
  \centering
  \includegraphics[width=0.45\textwidth]{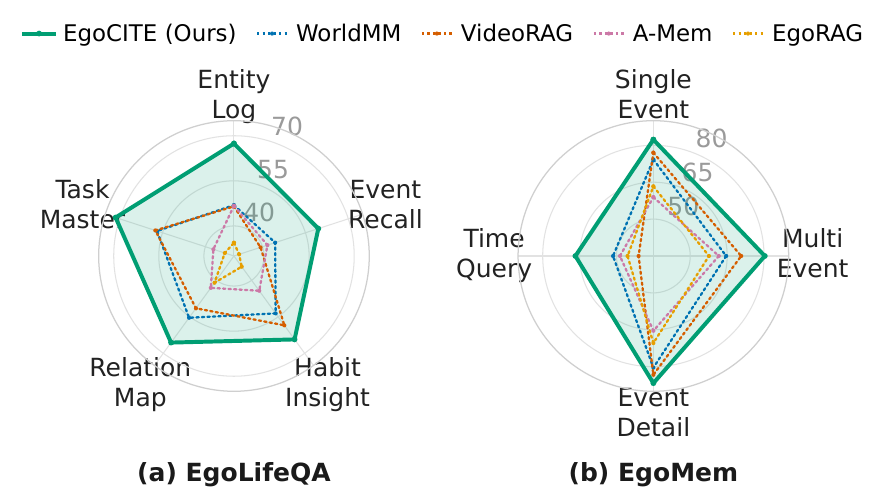}
  \caption{Per-category QA accuracy of agentic memory baselines on (a) EgoLifeQA~\citep{yang2025egolife} and (b) EgoMem~\citep{zheng2026lifedialbench} benchmarks.}
  \label{fig:radar-benchmark}
\end{wrapfigure}

Long-horizon egocentric memory aims to transform continuous first-person video and audio captured by wearable devices into a searchable record of daily life.
Such memory enables an agentic assistant to answer questions about experiences that might otherwise be forgotten: whom a user met, where an object was left, what was discussed, or what typically occurs in a particular situation.
This capability underpins emerging applications in memory augmentation, wearable augmented reality, and AI glasses that capture visual, audio, and social experiences~\citep{zulfikar2024memoro,paruchuri2025egotrigger,kim2026speechless,wang2026egoself}.
Realizing this vision requires organizing dense multimodal observations into a searchable index whose entries remain interpretable outside their local context, retrieving the right evidence from a long history according to a question's temporal intent, and reasoning over that evidence to produce an accurate answer.

Egocentric memory systems increasingly organize continuous experience into elaborate memory structures, such as knowledge graphs, semantic memories, and multi-scale hierarchies~\citep{yeo2026worldmm,sun2026egograph,yan2026ava}.
This mirrors agentic workflows in software engineering and computer use, where agents iteratively inspect and act on external workspaces~\citep{yang2024sweagent,xie2024osworld}.
However, our investigation exposes a more fundamental bottleneck when this paradigm is applied to long-horizon egocentric memory.
Memory entries are typically derived from short, independently generated video captions and speech transcripts, which often omit the local context needed to interpret people, objects, places, and elliptical utterances.
Once such context-poor content is indexed, neither a richer memory structure nor a stronger retrieval agent can reconstruct information that was never represented.
We therefore introduce \texttt{EgoScheme}, which uses local multimodal context to transform fragmentary video captions and speech transcripts into self-contained atomic memory indices.
\texttt{EgoIndex} then organizes disentangled and complementary action, activity, utterance, and conversation representations into searchable multi-view memory indices at multiple granularities.

Even with self-contained atomic memory indices, retrieval remains a second bottleneck.
Existing systems typically let an agent search captions or entities by semantic similarity and rerank the returned candidates~\citep{yang2025egolife,yeo2026worldmm}.
Yet semantic similarity captures what an event is about, not when or how often it occurred.
Questions over long-horizon egocentric memory frequently express temporal intent through qualifiers such as ``last,'' ``first,'' ``usually,'' or ``this morning.''
The most semantically similar event may therefore be neither the requested occurrence nor representative of a recurring habit.
We address this limitation with \texttt{EgoRetrv}, a time-aware retrieval framework that combines iterative semantic search with question-conditioned temporal relevance scoring and curation of retrieved evidence.
To separate targeted evidence discovery from cross-evidence reasoning, \texttt{EgoRetrv} adopts a dual-agent design: a drafting agent iteratively retrieves specific evidence using semantic and temporal queries, while a sampling agent reasons over the accumulated evidence in context to curate a coherent set aligned with the question's temporal intent.

We instantiate these ideas in \texttt{EgoCITE} (Fig.~\ref{fig:overview}), a long-horizon agentic memory framework for egocentric QA that couples context-augmented indexing with time-aware retrieval.
The design separates the responsibilities that existing systems entangle: local context is used at construction time to make evidence self-contained, while temporal and cross-event reasoning are applied at retrieval time.
Our contributions are as follows:
\begin{itemize}[leftmargin=*,itemsep=0pt,topsep=0pt,parsep=0pt]
    \item We identify two fundamental failures of current egocentric memory systems.
    (1) Fragmented and elliptical memory entries lack the context required for reliable indexing and agentic search.
    (2) Existing retrieval interfaces do not sufficiently integrate question context, particularly temporal intent, into memory search and selection. 
    \item We introduce \texttt{EgoCITE}, which builds context-augmented, multi-view atomic memory indices and uses dual-agent, time-aware retrieval to select and curate retrieved atomic memory indices according to temporal intent. 
    \item We evaluate \texttt{EgoCITE} on EgoLifeQA~\citep{yang2025egolife}, EgoMem~\citep{zheng2026lifedialbench}, and EgoR1-Bench~\citep{tian2025ego} using answer accuracy and target-event retrieval alignment. \texttt{EgoCITE} improves accuracy over agentic memory baselines by at least 4.4--14.2\% while achieving 36$\times$ lower cost than long-context LLM agents. Our project page and code are available at \url{https://egocite.github.io}.
\end{itemize}


\section{Background and Related Work}
\noindent
\textbf{Egocentric Lifelogging.}
Egocentric lifelogging captures continuous first-person experience to support memory recall and personal assistance.
Ego4D~\citep{grauman2022ego4d}, EgoSchema~\citep{mangalam2023egoschema}, HD-EPIC~\citep{perrett2025hd}, and Nymeria~\citep{ma2024nymeria} scaled egocentric video from short-clip action recognition to long-form temporal reasoning and multi-day multimodal recordings.
Memoro~\citep{zulfikar2024memoro}, EgoTrigger~\citep{paruchuri2025egotrigger}, SpeechLess~\citep{kim2026speechless}, and EgoSelf~\citep{wang2026egoself} build wearable and AI glasses systems for memory logging across visual, audio, and interaction modalities.
EgoLife~\citep{yang2025egolife}, LifeDialBench~\citep{zheng2026lifedialbench}, EgoMemReason~\citep{wang2026egomemreason}, and SuperMemory-VQA~\citep{alam2026supermemory} offer multi-day video-speech data and QA benchmarks for life assistant evaluation.
However, building memory from continuous multimodal sensor streams remains challenging: the streams are high-dimensional and cannot be indexed or queried directly.
Handling them poorly results in information loss, retrieval errors, and weak long-horizon recall.

\noindent
\textbf{Agentic Memory.}
Agentic memory gives LLM agents persistent, queryable knowledge beyond a single context window.
Prior work spans retrieval-augmented reasoning~\citep{lewis2020retrieval,yao2022react,asai2024self} and long-term memory systems~\citep{packer2023memgpt,xu2026mem,kang2025memory,rasmussen2025zep}.
In multimodal memory, VideoRAG~\citep{ren2026videorag}, AMEGO~\citep{goletto2024amego}, WorldMM~\citep{yeo2026worldmm}, EgoGraph~\citep{sun2026egograph}, and AVA~\citep{yan2026ava} enable long-context video QA through retrieval-augmented memory.
%
We summarize the workflow of existing multimodal memory systems in Fig.~\ref{fig:failure-case} through four key concepts.
\textit{(1) Memory entries.}
During construction, multimodal signals (i.e., video and audio) are transcribed by multimodal LLMs into natural-language captions, which serve as memory entries.
\textit{(2) Memory indices.}
Systems transform these entries into searchable indices through multi-granularity summarization~\citep{yang2025egolife,ren2026videorag}, entity extraction~\citep{yeo2026worldmm}, or keyword extraction~\citep{xu2026mem}.
\textit{(3) Memory structures.}
The resulting indices are stored in structures such as vector databases~\citep{yang2025egolife,ren2026videorag,xu2026mem} or knowledge graphs~\citep{yeo2026worldmm}.
\textit{(4) Retrieved evidence.}
At inference time, an LLM reformulates the user's question into search queries, retrieves candidate memory entries through keyword matching~\citep{xu2026mem}, vector search~\citep{ren2026videorag,yang2025egolife}, or graph traversal~\citep{yeo2026worldmm}, and reranks them by relevance.
The highest-ranked entries and their associated multimodal signals then become retrieved evidence for response generation.
%
%
In this work, we mainly investigate the role of high-quality memory indices in improving long-horizon egocentric memory question answering.


\section{Motivations}
\label{sec:motivations}

We evaluate existing long-horizon egocentric memory systems~\citep{yeo2026worldmm,ren2026videorag,xu2026mem,yang2025egolife} in Appx.~\ref{sec:appendix-baseline-failure} and identify two fundamental limitations.

\begin{figure*}[!t]
    \centering
    \includegraphics[width=\textwidth]{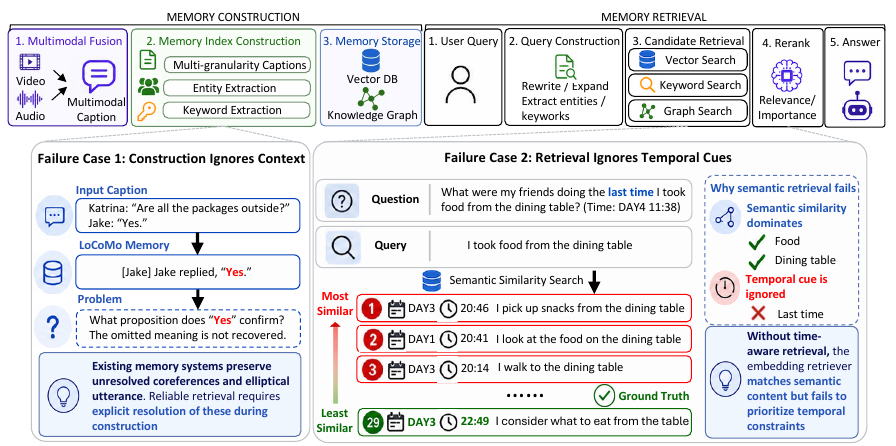}
    \caption{The workflow and failure cases of existing memory construction and retrieval approaches.}
    \label{fig:failure-case}
\end{figure*}

\noindent\textbf{Insight 1: Long-horizon egocentric agentic search is limited by what is indexed.}
Existing systems organize experience using increasingly elaborate memory structures, including vector databases, knowledge graphs, semantic memories, and visual memories~\citep{yang2025egolife,sun2026egograph,yeo2026worldmm}. Yet their memory indices are typically built directly from short, independently generated memory entries, such as captions and speech transcripts. These entries often omit the local context needed to identify people, objects, actions, and elliptical utterances, making relevant experiences difficult for an agent to retrieve.
Coreference resolution alone is insufficient because real-world conversations also contain ellipsis and omitted expressions~\citep{aralikatte2019ellipsis}. For example, in failure case 1 of Fig.~\ref{fig:failure-case}, Jake replies, ``\textit{yes},'' but the isolated entry does not specify what he confirms. In our analysis of WorldMM~\citep{yeo2026worldmm} and LoCoMo~\citep{maharana2024evaluating}, 10\% of WorldMM's extracted entities contain unresolved pronouns, and 17\% of LoCoMo memory entries contain unresolved verbatim quotes (Appx.~\ref{sec:appendix-failure-resolution}).
\emph{Context omitted from memory entries before indexing remains unavailable to downstream agentic search. Memory construction should therefore use local context to produce self-contained atomic memory indices.}

\noindent\textbf{Insight 2: Retrieval must model temporal intent in addition to semantic relevance.}
A second limitation is that egocentric memory retrieval often lacks \textit{time awareness}.
Memory recall questions are inherently temporal~\citep{gouveia2013footprint}, expressing recency (``last time''), habits (``usually''), or explicit time (``yesterday afternoon'').
More than 76\% of EgoLifeQA~\citep{yang2025egolife} and 48\% of EgoMem~\citep{zheng2026lifedialbench} questions contain temporal intent.
Existing systems nonetheless rank by semantic similarity and treat time as passive metadata~\citep{yang2025egolife, yeo2026worldmm, ren2026videorag, xu2026mem}, so the top-$k$ fills with semantically plausible but temporally wrong evidence. We show an example of this retrieval failure in Fig.~\ref{fig:failure-case}, where semantically similar memories are retrieved while the temporally correct evidence is ranked much lower.
This failure is reflected in EgoLifeQA, where accuracy on time-related questions is 9.8--10.3\% lower for WorldMM~\citep{yeo2026worldmm} and 13.3\% lower for VideoRAG~\citep{ren2026videorag} than on time-unrelated questions (Appx.~\ref{sec:appendix-failure-time-awareness}).
\textit{We therefore argue that temporal reasoning should be an explicit component of retrieval, rather than a post hoc filter over semantically retrieved memories.}

Making retrieval time-aware is not as simple as adding timestamps to the memory query for two reasons.
\textit{(1) Sentence embeddings ignore temporal intent.} Embeddings are optimized for semantic similarity rather than temporal ordering, so appending temporal expressions does not reliably retrieve the correct evidence.
\textit{(2) Temporal expressions are often ambiguous.} Phrases such as ``at noon'' or ``at breakfast'' refer to approximate time ranges rather than exact timestamps, making hard temporal filtering prone to missing relevant memories~\citep{sun2026egograph}.


\section{Methodology}
\begin{figure*}[t!]
    \centering
    \includegraphics[width=\textwidth]{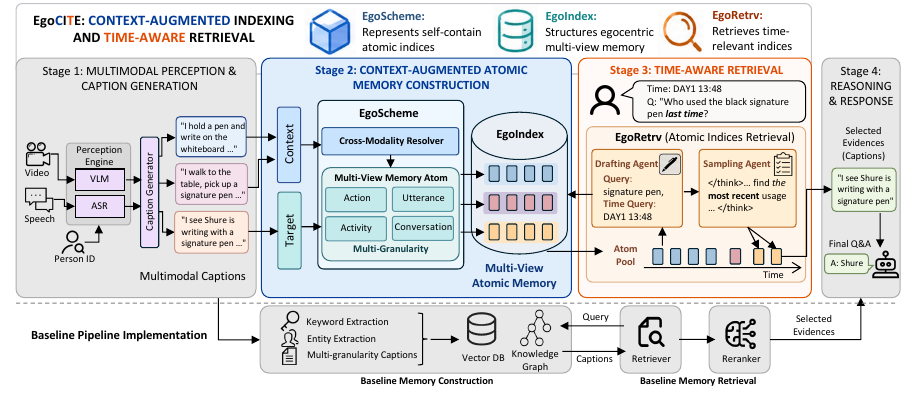}
    \caption{Overview of the pipeline: (1) Multimodal perception \& caption generation, (2) Context-augmented memory indexing, (3) Time-aware atomic memory index retrieval, and (4) Reasoning \& response.}
    \label{fig:overview}
\end{figure*}

Together, these findings motivate \texttt{EgoCITE}, a four-stage pipeline for long-horizon egocentric memory (Fig.~\ref{fig:overview}). Stages 1 and 4 follow the standard perception and response architecture of existing egocentric memory systems, whereas our contributions lie in Stages 2 and 3.

\noindent\textbf{Stage 1: Multimodal Perception and Caption Generation.} We convert egocentric video and audio into dense visual captions and speech transcripts, resolve speaker and participant identities, and fuse these outputs into multimodal captions that serve as raw memory entries for Stage 2 (see implementation details in Appx.~\ref{sec:appendix-implementation-perception}).
\noindent\textbf{Stage 2: Context-Augmented Memory Indexing.}
Each raw memory entry is augmented with a context window of neighboring entries.
Within this window, \texttt{EgoScheme} resolves coreferences and ellipsis to produce self-contained atomic memory indices.
\texttt{EgoIndex} organizes these indices into four complementary views---actions, activities, utterances, and conversations---forming a searchable multi-view memory index (Sec.~\ref{sec:context-augment-construction}).
\noindent\textbf{Stage 3: Time-Aware Retrieval.} Given a user question, \texttt{EgoRetrv} retrieves atomic memory indices using two agents. A \textit{drafting} agent performs multi-round retrieval over \texttt{EgoIndex} using semantic similarity and \textit{time queries}, accumulating candidates in an \textit{index pool}. A \textit{sampling} agent then curates this pool with explicit awareness of temporal qualifiers such as ``last'', ``first'', and ``usually'' in the question (Sec.~\ref{sec:time-aware-evidence-retrieval}).
\noindent\textbf{Stage 4: Reasoning and Response.} The response agent receives the curated atomic memory indices and maps their timestamps back to multimodal captions. It then examines the selected captions and produces the final multiple-choice answer (see implementation details in Appx.~\ref{sec:appendix-implementation-response}).

\subsection{Context-Augmented Memory Indexing}
\label{sec:context-augment-construction}

Context-augmented memory indexing addresses the first limitation identified in Sec.~\ref{sec:motivations}: raw memory entries often lack the context required for reliable agentic search. We first define the multi-view, multi-granularity atomic memory indices exposed to the retrieval agent through \texttt{EgoIndex}, then describe how \texttt{EgoScheme} constructs these self-contained indices from context-poor memory entries.

\noindent \textbf{Design 1: \texttt{EgoIndex}---Multi-View Egocentric Memory Indexing.}
Existing systems typically construct memory indices directly from captions generated by multimodal LLMs, yielding a single, undifferentiated view of each observation~\citep{yang2025egolife,ren2026videorag}.
A single caption may combine short actions, extended activities, individual utterances, and broader conversations, even though a question may target only one interaction type or level of abstraction. \texttt{EgoIndex} therefore separates experience along two axes: physical behavior versus spoken interaction, and fine- versus coarse-grained semantics. Inspired by recent egocentric vision research~\citep{ma2024nymeria, yeo2026worldmm}, this organization yields four complementary memory \textit{views}: \textit{actions}, \textit{activities}, \textit{utterances}, and \textit{conversations}.
Actions and activities represent physical behavior at fine and coarse granularities, respectively, whereas utterances and conversations represent spoken interaction at fine and coarse granularities. Actions describe short-term behaviors, including hand-object interactions, meaningful gestures, and movements. Activities capture higher-level, goal-oriented interactions. Utterances record individual speech acts with self-contained content, whereas conversations summarize broader topic exchanges among participants. Each view-specific representation is an \textit{atomic memory index} derived from one or more raw memory entries.
Formally, each atomic memory index $m_i$ is associated with a timestamp $\tau_i$ inherited from its source memory entry or entries. We define \texttt{EgoIndex} as a collection of timestamped atomic memory indices,
{\small
\begin{equation*}
\mathcal{M}=\bigcup_{v\in\mathcal{V}}\mathcal{M}_v=\{(\tau_i,m_i)\}_{i=1}^{|\mathcal{M}|}.
\end{equation*}
}
where the four views are $\mathcal{V}=\{\textsc{Action},\textsc{Utterance},\textsc{Activity},\textsc{Conversation}\}$.
\texttt{EgoIndex} offers two advantages:
(1) \textit{Multi-granularity retrieval.} Fine- and coarse-grained atomic memory indices support queries at different abstraction levels.
(2) \textit{Separation of interaction types.} Physical behavior and spoken interaction capture complementary aspects of an experience.
\begin{table*}[t]
\centering
\scriptsize
\setlength{\tabcolsep}{3pt}
\begin{tabular}{
    p{0.08\textwidth}
    p{0.08\textwidth}
    p{0.4\textwidth}
    p{0.38\textwidth}
}
\toprule
\textbf{View} & \textbf{Component} & \textbf{Specification} & \textbf{Example} \\
\midrule

\multirow{5}{*}{Action}
& \multirow{2}{*}{HOI}
& Require a named \textbf{person}, a specific \textbf{object}, and \textbf{direct physical handling}.
& \multirow{2}{*}{``\textbf{I fasten} the \textbf{shelf bracket} with \textbf{a screwdriver}.''} \\
\cline{2-4}

&  \multirow{1}{*}{Gesture}
& Require a named \textbf{person} and a \textbf{meaningful expression}.
& \multirow{1}{*}{``\textbf{Jake laughs loudly}.''} \\
\cline{2-4}

& \multirow{2}{*}{Movement}
& Require a named \textbf{person} and a named \textbf{destination} that the person reaches.
& \multirow{2}{*}{``\textbf{Shure} enters the \textbf{living room}.''} \\
\midrule

\multirow{2}{*}{Utterance}
&
& Require a named \textbf{speaker}, a \textbf{speech act}, and \textbf{self-contained content with resolved pronouns and ellipsis}.
& ``\textbf{Jake tells} Tasha that \textbf{dinner is ready}.'' ``\textbf{Jake confirms} that \textbf{all the packages are outside}.'' \\
\midrule

\multirow{2}{*}{Activity}
&
& Require an \textbf{immediate goal}, \textbf{specific participants, an object, and a location}.
& \multirow{2}{*}{``\textbf{Alice, Tasha, and I} \textbf{prepare dinner} in \textbf{the kitchen}.''} \\
\midrule

\multirow{2}{*}{Conversation}
&
& Require \textbf{participants} and \textbf{specific, coherent topic} across captions.
& ``\textbf{Jake and I} debate \textbf{which model to use for the demo video}.'' \\
\bottomrule
\end{tabular}

\caption{Key specifications and examples for \texttt{EgoScheme}.}
\label{tab:egoscheme-prompt-views}
\end{table*}

\noindent \textbf{Design 2: \texttt{EgoScheme}---Context-Augmented Atomic Memory Index Construction.}
To construct self-contained atomic memory indices for \texttt{EgoIndex}, \texttt{EgoScheme} defines a structured specification for each view, as shown in Tab.~\ref{tab:egoscheme-prompt-views}.
All four views follow two principles: (1) \textit{Human-centered}: each atomic memory index is anchored to a specific person; and
(2) \textit{Decoupled}: each index captures one coherent behavior, activity, utterance, or conversational topic, preventing compounded semantics from confusing similarity-based retrieval. When the available context does not support a unique resolution, \texttt{EgoScheme} preserves the original expression rather than introducing unsupported details.

For (a) action, we divide user behavior into hand-object interactions (HOIs), gestures, and movements. An HOI specifies the person, object, and direct physical interaction. A gesture captures a meaningful expression. A movement specifies a user's movement toward a named destination. We demonstrate examples in Tab.~\ref{tab:egoscheme-prompt-views}.
For (b) utterance, each atomic memory index specifies the speaker, speech act, and self-contained content with resolved pronouns and ellipsis. For example, ``Jake confirms that all the packages are outside'' identifies Jake as the speaker, ``confirms'' as the speech act, and ``all the packages are outside'' as the complete propositional content. 
To extract (a) actions and (b) utterances, we set the context segment length to $T$ ($T=5$ min by default), such that the context of each caption $C_t$ consists of all captions from the preceding $T$ minutes, denoted by $[C_{t-T}, \ldots, C_t]$. For each current caption $C_t$, we prompt an LLM to generate structured atomic memory indices following the specifications above, conditioned on the entire context window. The extracted index $m_i$ inherits timestamp $\tau_i=t$ from the source caption.

For (c) activity, each atomic memory index summarizes one immediate goal grounded in its participants, object, and location. 
For (d) conversation, each atomic memory index names all participants and captures one coherent topic across captions. We demonstrate examples in Tab.~\ref{tab:egoscheme-prompt-views}.
To extract (c) and (d), we partition the multimodal captions into disjoint $T'$-minute windows, denoted by $[C_{t-T'},\ldots,C_{t}]$, with $T'=30$ minutes by default following WorldMM~\citep{yeo2026worldmm}.
We prompt an LLM to segment each window into activity or conversation atomic memory indices $m_i$ following \texttt{EgoScheme}, with each index assigned a specific temporal interval $[t_i^{\mathrm{start}},t_i^{\mathrm{end}}]$. The extracted index $m_i$ inherits the start time of its source caption segment, i.e., $\tau_i=t_i^{\mathrm{start}}$.
To ensure that each memory index captures a single coherent behavior, we further apply a disentangling procedure that separates unrelated content joined by connectives such as ``and'' or ``while.''
To reconcile indices across adjacent windows, we identify temporally continuous memory indices using sentence-embedding similarity and merge them with an additional LLM call (see Appx.~\ref{sec:appendix-implementation-construction} for implementation details and Appx.~\ref{sec:appendix-examples} for examples).

\subsection{Time-Aware Atomic Memory Index Retrieval}
\label{sec:time-aware-evidence-retrieval}
Motivated by Insight 2 in Sec.~\ref{sec:motivations}, \texttt{EgoRetrv} makes a question's temporal intent an explicit retrieval signal. It combines question-conditioned temporal relevance scoring with a dual-agent design:
(1) A \textit{drafting} agent issues multi-round semantic, view, and \textit{time} queries to \texttt{EgoIndex} and progressively accumulates specific atomic memory indices in a persistent working memory named the \textit{index pool}, prioritizing retrieval recall.
(2) A \textit{sampling} agent reasons over the accumulated indices in context and curates a coherent evidence set aligned with the question's temporal intent, improving retrieval precision by selecting temporally relevant indices and removing redundant ones.

\noindent \textbf{Design 3: Drafting agent with time query.}
At each retrieval round, given a question $Q$, the drafting agent generates a structured query set $\mathcal{Q}=\{Q_s,Q_v,Q_t\}$, where $Q_s$ is a semantic query, $Q_v$ specifies the memory view to retrieve, and $Q_t$ is an optional \textit{temporal query}.
Using a sentence-embedding model $f$, we compute the semantic relevance of each candidate memory index $m_i\in\mathcal{M}_{Q_v}$ as $S_i=\operatorname{sim}\left(f(Q_s),f(m_i)\right)$.

The temporal query $Q_t=[t_{\mathrm{start}},t_{\mathrm{end}}]$ represents the drafting agent's estimated time range for the queried event,
inferred from the user question using commonsense temporal knowledge and atomic memory indices accumulated in prior rounds.
A point estimate is represented by $t_{\mathrm{start}}=t_{\mathrm{end}}$. For example, a ``last time'' qualifier with a reference time of \texttt{DAY1 12:00} may yield $Q_t=[\texttt{DAY1 12:00},\ \texttt{DAY1 12:00}]$, whereas ``this morning'' may yield $Q_t=[\texttt{DAY1 06:00},\ \texttt{DAY1 12:00}]$.
The agent may update $Q_t$ across retrieval rounds as the retrieved atomic memory indices refine its temporal belief.
We derive the temporal relevance score $R_i$ directly from the temporal query $Q_t=[t_{\mathrm{start}},t_{\mathrm{end}}]$. After mapping all timestamps onto a continuous timeline
with hour as the unit, we compute
{\small
\begin{equation}
R_i=
\begin{cases}
1, & t_{\mathrm{start}} \leq \tau_i \leq t_{\mathrm{end}}, \\[1pt]
\lambda^{t_{\mathrm{start}}-\tau_i},
& \tau_i < t_{\mathrm{start}}, \\[1pt]
\lambda^{\tau_i-t_{\mathrm{end}}},
& \tau_i > t_{\mathrm{end}}.
\end{cases},
\label{eq:time-decay}
\end{equation}
}
where $0<\lambda<1$ controls the time-decay factor. We select the default $\lambda=0.99$ based on ablation study in Sec.~\ref{sec:experiments}.
Candidates within $Q_t$ receive $R_i=1$, whereas candidates outside the interval are exponentially downweighted according to their hourly distance from the nearest boundary of $Q_t$.
If no temporal query $Q_t$ is provided, we set $R_i=1$ by default.
The final retrieval score combines semantic and temporal relevance multiplicatively:
{\small
\begin{equation*}
\operatorname{score}(m_i\mid Q_s,Q_v,Q_t)=S_iR_i.
\end{equation*}
}
The top-$k$ timestamped atomic memory indices are selected according to the final score:
{\small
\begin{equation*}
\mathcal{K}=
\operatorname{TopK}_{(\tau_i,m_i)\in\mathcal{M}_{Q_v}}
\operatorname{score}(m_i\mid Q_s,Q_v,Q_t),
\end{equation*}
}
where $\mathcal{K} \in \mathcal{M}_{Q_v}$ contains the retrieved \texttt{(timestamp, index)} pairs from the specified view. The index pool $\mathcal{P}$ is updated as $\mathcal{P}\leftarrow\mathcal{P}\cup\mathcal{K}$.
Implementation details are presented in Appx.~\ref{sec:appendix-implementation-drafting-agent} and~\ref{sec:appendix-examples}.

Together, these mechanisms address the two challenges identified in Insight 2. (1) Modeling temporal relevance $R_i$ outside the embedding space avoids the temporal insensitivity of sentence embeddings. (2) Soft decay via $\lambda$ accommodates ambiguous temporal expressions by assigning nearby atomic memory indices non-zero scores. In contrast, hard temporal cutoffs~\citep{sun2026egograph} discard all memory indices outside a fixed window and are brittle to reasoning errors and linguistic ambiguity.

\noindent \textbf{Design 4: Sampling agent with time-aware curation.}
The time query biases retrieval toward a predicted window but lacks a global view of the collected atomic memory indices.
We therefore decouple retrieval and curation: the drafting agent collects high-recall candidates, while the sampling agent reviews them and selects the temporally correct atomic memory indices based on time cues in the question statement.
In practice, the sampling agent reads the timestamped atomic memory indices from the \textit{index pool} $\mathcal{P}$ and orders them chronologically. It then applies the curation $\mathcal{C}(Q,\mathcal{P})$ to select the most relevant indices based on the temporal cues in the question $Q$.
For example, for recency-oriented questions containing qualifiers such as ``last'' or ``most recent,'' the sampling agent prioritizes atomic memory indices with the latest timestamps while retaining related indices needed to distinguish among answer choices. For habitual cues such as ``usually'' or ``habit,'' it selects a temporally diverse set of relevant atomic memory indices spanning multiple days to capture recurring behavior rather than a single event (see implementation details in Appx.~\ref{sec:appendix-implementation-sampling-agent}).
This extends the recency-only selection algorithm in EgoRAG~\citep{yang2025egolife} to questions that require multiple atomic memory indices to answer, such as habits, coreferences, and specific time ranges. This also outperforms semantic memory in WorldMM~\citep{yeo2026worldmm}, where habitual questions are poorly summarized during construction time due to imperfect memory consolidations (see Sec.~\ref{sec:experiments}).


\section{Experiments}
\label{sec:experiments}
\noindent\textbf{Benchmarks.}
We evaluate \texttt{EgoCITE} on three egocentric memory benchmarks on EgoLife~\citep{yang2025egolife}, including EgoLifeQA~\citep{yang2025egolife}, EgoMem~\citep{zheng2026lifedialbench}, and EgoR1-Bench~\citep{tian2025ego}. 
For benchmarks, we use question statements, answer choices, and question time as the input context, and use ground-truth answers and ground-truth timestamps that requires to answer the question for evaluation (See details in Appx.~\ref{sec:appendix-evaluation}). 

\noindent\textbf{Baselines.}
We compare \texttt{EgoCITE} with long-context LLM agents Gemini-3.1-Pro~agent~\citep{team2023gemini} and GPT-5.4~agent~\citep{openai2026gpt54thinking}, agentic memory baselines, including EgoRAG~\citep{yang2025egolife}, A-MEM~\citep{xu2026mem}, VideoRAG~\citep{ren2026videorag}, and WorldMM~\citep{yeo2026worldmm}. Most baselines are evaluated using GPT-5.4 for retrieval agents and WorldMM is evaluated with Qwen3.6-27B-FP8~\citep{qwen3.6} additionally. We evaluate \texttt{EgoCITE} with both GPT-5.4 and Qwen3.6-27B-FP8 at default 5-round retrieval and 15 memory indices to align with WorldMM. Additional details on dataset preprocessing, metrics, baseline implementations, model configurations, and hardware are provided in Appx.~\ref{sec:appendix-implementation} and \ref{sec:appendix-evaluation}.

\noindent\textbf{Metrics.}
We use multiple-choice accuracy as the main metric. We use input token numbers, retrieval latency, and normalized cost as efficiency metrics. \textit{Normalized Cost} measures the normalized API token cost of GPT-5.4~\citep{openai2026pricing}. Short- and long-context models use normalized input/output costs of $1\times$/$6\times$ and $2\times$/$9\times$, respectively. \texttt{EgoCITE-GPT} and other agentic memory baselines use short-context pricing and the long-context GPT agent uses long-context pricing. The normalized cost is computed as the weighted sum of input and output token costs (details in Appx.~\ref{sec:appendix-metrics}). 

\noindent \textbf{Main Results.}
We report the results in Tab.~\ref{tab:main-result}. \texttt{EgoCITE-GPT} is the strongest baseline, outperforming the long-context LLM-agent baselines by 3.6--8.9\% on EgoLifeQA, 0.9--4.1\% on EgoMem, and 2.6--4.7\% on EgoR1-Bench. 
Compared with agentic memory baselines, it achieves average gains of at least 14.2\%, 4.4\%, and 9.0\% on the three benchmarks, respectively.
\texttt{EgoCITE-Qwen} trails \texttt{EgoCITE-GPT} by 1--3\% average accuracy while still outperforming all baselines. 
Long-context LLM agents consistently outperform existing agentic memory systems.
Nevertheless, \texttt{EgoCITE} surpasses these long-context LLM agents while requiring 23$\times$ fewer input tokens, demonstrating that accurate long-term memory retrieval does not require million-token contexts.

\noindent \textbf{Retrieval Hit Rate.}
We report the retrieval hit rate over the agent's retrieved memory indices, with a maximum retrieval budget of 15 memory indices for all baselines before question answering (Table~\ref{tab:hit-rate}). Both \texttt{EgoCITE} implementations substantially outperform existing retrieval-based memory methods across all three benchmarks. \texttt{EgoCITE-GPT} achieves the highest hit rates of 49.6\%, 89.6\%, and 62.7\% on EgoLifeQA, EgoMem, and EgoR1, respectively, improving over the best baseline by up to 15.1\%, 17.0\%, and 22.7\%. These results indicate that \texttt{EgoCITE}'s memory representation and time-aware retrieval enable substantially more accurate retrieval, regardless of the underlying model's capability.

\begin{table*}[t]
\centering
\scriptsize
\caption{Baseline accuracy (\%) and number of input tokens on three benchmarks. Bold, underlined, and italicized values denote the best, second-best, and third-best results, respectively.}
\label{tab:main-result}
\setlength{\tabcolsep}{1pt}
\begin{tabular}{lllcccccccccccc}
\toprule
\multirow{3}{*}{\textbf{Method}} &
\multirow{3}{*}{\textbf{Model}} &
\multirow{3}{*}{\shortstack{\textbf{Input}\\\textbf{Token}}} &
\multicolumn{6}{c}{\textbf{EgoLifeQA}~{\scriptsize\citep{yang2025egolife}}} &
\multicolumn{5}{c}{\textbf{EgoMem}~{\scriptsize\citep{zheng2026lifedialbench}}} &
\multirow{2}{*}{\shortstack{\textbf{EgoR1-Bench}\\
{\scriptsize\citep{tian2025ego}}}} \\
\cmidrule(lr){4-9}
\cmidrule(lr){10-14}
&
&
&
\textbf{Ent.} &
\textbf{Evt.} &
\textbf{Hab.} &
\textbf{Rel.} &
\textbf{Task} &
\textbf{Avg.} &
\textbf{Sgl.} &
\textbf{Mul.} &
\textbf{Det.} &
\textbf{Time} &
\textbf{Avg.} &
\\
\midrule

GPT-5.4 agent~\citep{openai2026gpt54thinking}
& --
& 783k
& 56.2
& 48.1
& \underline{63.9}
& 57.3
& \textit{69.7}
& 54.9
& 80.6 
& \textbf{83.8} 
& 84.6
& 49.5
& 75.1 
& 71.0 \\

Gemini-3.1-Pro agent~\citep{team2023gemini}
& --
& 804k
& \textit{61.1}
& \textit{56.7}
& \textit{59.8}
& \textit{62.8}
& \underline{70.1}
& \textit{60.2}
& \textit{81.0}
& \underline{81.2}
& 81.7
& \textbf{68.5}
& \textit{78.3}
& \textit{72.7}
\\

\midrule
EgoRAG~{\scriptsize\citep{yang2025egolife}}
& GPT
& 16k
& 34.4
& 31.8
& 34.4
& 41.0
& 33.2
& 34.3
& 63.3
& 57.6
& 70.4
& 45.5
& 59.5
& 49.5
\\

A-MEM~{\scriptsize\citep{xu2026mem}}
& GPT
& 6k
& 46.7
& 41.7
& 44.3
& 43.1
& 37.2
& 42.9
& 58.9
& 61.6
& 65.4
& 48.6
& 58.8
& 52.0
\\

VideoRAG~{\scriptsize\citep{ren2026videorag}}
& GPT
& 21k
& 46.5
& 39.3
& 58.5
& 51.5
& 57.5
& 46.5
& 77.0
& 70.7
& 83.3
& 41.0
& 68.6
& 64.7
\\

WorldMM-Qwen~{\scriptsize\citep{yeo2026worldmm}}
& Qwen
& 87k
& 46.1
& 41.3
& 49.5
& 49.2
& 54.6
& 46.4
& 77.8
& 77.3
& \textit{84.6}
& 58.1
& 74.8
& 66.3
\\

WorldMM-GPT~{\scriptsize\citep{yeo2026worldmm}}
& GPT
& 134k
& 46.9
& 44.4
& 53.6
& 55.4
& 56.9
& 49.6
& 74.2
& 64.6
& 80.4
& 51.4
& 68.1
& 64.0
\\

\midrule

\textbf{\texttt{EgoCITE-Qwen} (Ours)}
& Qwen
& 34k
& \underline{62.7}
& \textbf{59.5}
& 58.8
& \underline{63.6}
& 67.0
& \underline{61.5}
& \textbf{83.5}
& 78.2
& \textbf{88.3}
& \underline{67.6}
& \textbf{79.9}
& \underline{73.0}
\\

\textbf{\texttt{EgoCITE-GPT} (Ours)}
& GPT
& 32k
& \textbf{67.4}
& \textbf{59.5}
& \textbf{64.3}
& \textbf{65.6}
& \textbf{71.4}
& \textbf{63.8}
& \underline{82.3}
& \textit{80.3}
& \underline{86.7}
& \textit{66.7}
& \underline{79.2}
& \textbf{75.3}
\\

\bottomrule
\end{tabular}
\end{table*}

\begin{table*}[t]
\centering

\begin{minipage}[t]{0.55\textwidth}
\centering
\captionof{table}{Retrieval memory index hit rate (\%). Green color for improvements compared to the best baseline.}
\label{tab:hit-rate}

\scriptsize
\setlength{\tabcolsep}{2pt}
\begin{tabular}{lccc}
\toprule
\textbf{Method} & \textbf{EgoLifeQA} & \textbf{EgoMem} & \textbf{EgoR1} \\
\midrule
EgoRAG{~\citep{yang2025egolife}}
    & 4.3  & 41.1 & 7.6 \\
A-Mem{~\citep{xu2026mem}}
    & 15.4 & 41.4 & 24.7 \\
VideoRAG{~\citep{ren2026videorag}}
    & 2.0  & 17.4 & 6.0 \\
WorldMM-Qwen{~\citep{yeo2026worldmm}}
    & 31.0 & \textit{72.6} & 38.3 \\
WorldMM-GPT{~\citep{yeo2026worldmm}}
    & \textit{34.5} & 66.9 & \textit{40.0} \\
\midrule
\textbf{\texttt{EgoCITE-Qwen} (Ours)}
    & \underline{43.4}\textcolor{green!60!black}{\tiny(+8.9)}
    & \underline{86.8}\textcolor{green!60!black}{\tiny(+14.2)}
    & \underline{59.0}\textcolor{green!60!black}{\tiny(+19.0)} \\
\textbf{\texttt{EgoCITE-GPT} (Ours)}
    & \textbf{49.6}\textcolor{green!60!black}{\tiny(+15.1)}
    & \textbf{89.6}\textcolor{green!60!black}{\tiny(+17.0)}
    & \textbf{62.7}\textcolor{green!60!black}{\tiny(+22.7)} \\
\bottomrule
\end{tabular}
\end{minipage}
\hfill
\begin{minipage}[t]{0.42\textwidth}
\centering
\captionof{table}{Retrieval-round ablation for
\texttt{EgoCITE-GPT} on
EgoLifeQA.}
\label{tab:retrieval-depth}

\scriptsize
\setlength{\tabcolsep}{2pt}

\begin{tabular}{lccc}
\toprule
\textbf{Metric} &
\textbf{1-Round} &
\textbf{3-Round} &
\textbf{5-Round} \\
\midrule
Hit rate (\%) & 36.5 & 48.6 & 49.6 \\
Average accuracy (\%)  & 58.7 & 62.6 & 63.8 \\
\midrule
EntityLog    & 61.9 & 66.1 & 67.4 \\
EventRecall  & 54.7 & 58.6 & 59.5 \\
HabitInsight & 60.1 & 60.1 & 64.3 \\
RelationMap  & 62.6 & 65.4 & 65.6 \\
TaskMaster   & 61.0 & 71.0 & 71.4 \\
\bottomrule
\end{tabular}
\end{minipage}
\end{table*}

\begin{figure}[t]
\centering
\begin{minipage}[t]{0.45\linewidth}
    \centering
    \includegraphics[width=\linewidth]{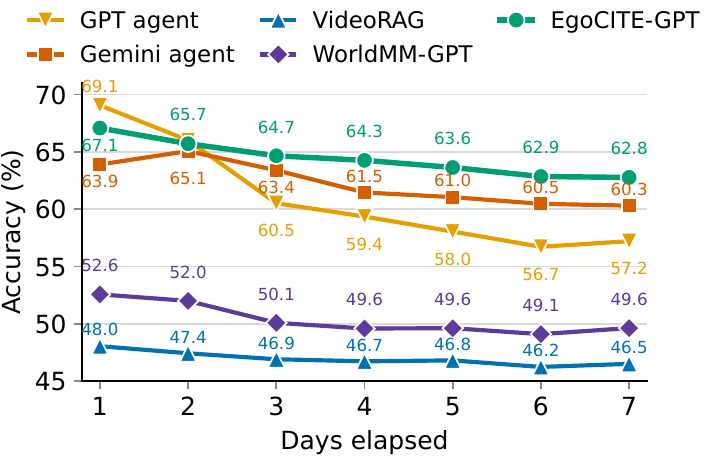}
    \captionof{figure}{Memory scaling: EgoLifeQA accuracy over increasing memory horizon. Differences are compared to DAY1.}
    \label{fig:cumulative-accuracy}
\end{minipage}
\hfill
\begin{minipage}[t]{0.45\linewidth}
    \centering
    \includegraphics[width=\linewidth]{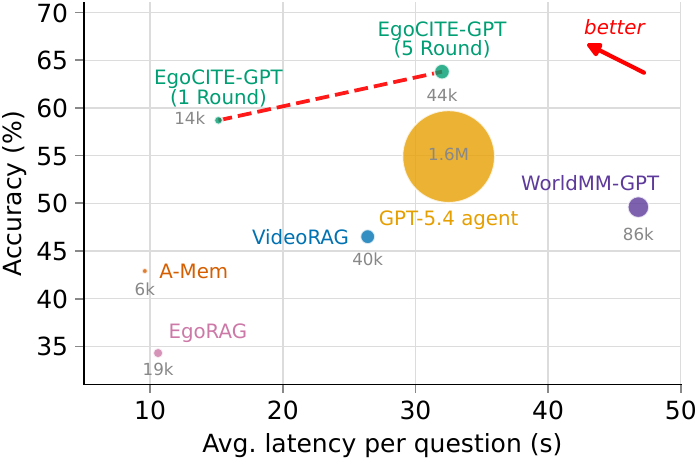}
    \captionof{figure}{EgoLifeQA accuracy, latency, and normalized cost trade-offs. Gray annotations and marker size indicate normalized cost.}
    \label{fig:efficiency}
\end{minipage}
\end{figure}
\begin{wraptable}{r}{0.41\textwidth}
\centering
\scriptsize
\setlength{\tabcolsep}{1pt}
\begin{tabular}{lc}
\toprule
\textbf{Method} & \textbf{Accuracy (\%)} \\
\midrule
EgoRAG~\citep{yang2025egolife} & 34.8 \\
A-Mem~\citep{xu2026mem} & 42.2 \\
VideoRAG~\citep{ren2026videorag} & 43.4 \\
WorldMM-Qwen~\citep{yeo2026worldmm} & 43.8 \\
WorldMM-GPT~\citep{yeo2026worldmm} & 47.1 \\
\midrule
GPT-5.4 agent~\citep{openai2026gpt54thinking} & 51.6 \\
Gemini-3.1-Pro agent~\citep{team2023gemini} & \textit{58.7} \\
\midrule
\textbf{\texttt{EgoCITE-Qwen} (Ours)}
    & \underline{60.9} \\
\textbf{\texttt{EgoCITE-GPT} (Ours)}
    & \textbf{62.6} \\
\bottomrule
\end{tabular}
\caption{Time-aware questions.}
\label{tab:temporal-aware-accuracy}
\end{wraptable}
\noindent \textbf{Temporal-Aware Questions.} 
We demonstrate the accuracy of EgoLifeQA with temporal-aware questions (see Appx.~\ref{sec:appendix-dataset-benchmark}) in Tab.~\ref{tab:time-relevant-questions}.
\texttt{EgoCITE-GPT} achieves the highest accuracy at 62.6\%, followed by \texttt{EgoCITE-Qwen} at 60.9\%. 
Both variants outperform the strongest long-context LLM baseline Gemini-3.1-Pro~agent by 3.9\% and 2.2\%. 
Compared with the strongest agentic memory WorldMM-GPT, they improve accuracy by 15.5\% and 13.8\%.
These results demonstrate that \texttt{EgoCITE} more effectively captures temporal intent than both long-context LLM agents and existing agentic memory baselines.

\noindent \textbf{Habits and Multiple Evidences.}
Habitual and multi-evidence questions require retrieving information across multiple timestamps. 
We report the results on habitual (Hab.) questions in EgoLifeQA and multi-evidence (Mul.) questions in EgoMem in Tab.~\ref{tab:main-result}. 
\texttt{EgoCITE-GPT} ranks first and third behind the long-context GPT-5.4~agent. It outperforms agentic memory baselines by 5.8--29.9\% and 3.0--22.7\% on two categories, while requiring more than 24$\times$ fewer input tokens than long-context LLM agents. These results demonstrate that \texttt{EgoRetrv} effectively retrieves information across long temporal horizons without million-token contexts.

\noindent \textbf{Memory Scaling.}
Fig.~\ref{fig:cumulative-accuracy} reports cumulative accuracy on EgoLifeQA as the memory horizon increases from DAY1 to DAY7. \texttt{EgoCITE} achieves the highest accuracy from DAY3 onward and shows only a 4.8\% decline from DAY1 to DAY7, compared with an 11.9\% drop for GPT-5.4~agent.
Gemini-3.1-Pro~agent, WorldMM-GPT, and VideoRAG maintain lower accuracy throughout. These results show that \texttt{EgoCITE}'s structured multi-view memory and time-aware retrieval scale more effectively than both long-context reasoning and existing retrieval-based approaches.

\noindent \textbf{Retrieval Rounds.}
We ablate retrieval depth in \texttt{EgoCITE-GPT} using 1-, 3-, and 5-round variants. Increasing retrieval depth improves hit rate and QA accuracy, with gains of 13.1\% and 5.1\% from 1 to 5 rounds (Tab.~\ref{tab:retrieval-depth}). Most gains are achieved within the first three rounds and later rounds provide only modest improvements, demonstrating the benefit of progressive multi-round retrieval.

\noindent \textbf{Retrieval Efficiency.}
Since existing baselines differ in retrieval workflows, memory representations, and modalities, we compare their trade-offs among accuracy, retrieval latency, and normalized cost.
5-round variant achieves the highest accuracy with 2$\times$ lower normalized cost and 15s lower latency than WorldMM, thanks to atomic memory index retrieval that reduces retrieved context by 5.7$\times$.
5-round variant also outperforms GPT-5.4 agent by 8.9\% accuracy and 36$\times$ cost savings with comparable latency. 
Five-round retrieval improves accuracy by 5.1\% over one round, with 2$\times$ higher latency and 3.2$\times$ higher normalized cost. Overall, \texttt{EgoCITE} is Pareto-optimal in accuracy, latency, and cost.

\noindent\textbf{Time-Decay Factor.}
We ablate the temporal decay factor $\lambda$ of \texttt{EgoCITE-GPT} on a 100-question validation subset of EgoR1-Bench in Tab.~\ref{tab:decay-lambda-ablation}. Performance improves as $\lambda$ increases from 0.97 to 0.99, peaking at 81.0\% accuracy and 67.0\% hit rate. Increasing $\lambda$ further to 0.995 reduces both metrics, indicating that $\lambda=0.99$ best balances semantic similarity and temporal relevance.

\noindent \textbf{Component Ablation.}
We perform component ablations on \texttt{EgoCITE-GPT} and report the results on EgoR1-Bench in Tab.~\ref{tab:component-ablation}. We incrementally add \texttt{EgoCITE} components on EgoRAG-like caption RAG. Starting from caption RAG, adding \texttt{EgoIndex}, \texttt{EgoScheme}, and \texttt{EgoRetrv} progressively improves accuracy from 61.3\% to 69.7\%, 71.3\%, and 75.3\%, and hit rate from 40.7\% to 54.3\%, 59.3\%, and 62.7\%, demonstrating the contribution of each component.

\noindent \textbf{Video Caption.} Following WorldMM~\citep{yeo2026worldmm}, we use the human-annotated dense captions and speech transcripts from EgoLife as the memory source for fair comparison with baselines. For realistic evaluation, we replace dense captions with narrations generated by visual-language model (VLM) Gemini-3-Flash~\citep{team2023gemini} and Gemma-4-31B~\citep{team2026gemma} (Appx.~\ref{sec:appendix-implementation-perception}). We use EgoLifeQA as benchmark and Qwen3.6-27B-FP8 as retrieval agent. Results show Gemini-3-Flash and Gemma-4-31B achieve 60.3\% and 57.1\% accuracy, only 1.2\% and 4.4\% lower than human-annotated dense captions (see details in Sec.~\ref{sec:appendix-additional-results}). These results demonstrate that \texttt{EgoCITE} generalizes well to realistic VLM-generated narrations while maintaining competitive long-term memory performance.


\section{Conclusion}
We presented \texttt{EgoCITE}, a memory system for long-horizon egocentric life assistants. The system is built around a simple premise: a memory index is only as useful as the entries it organizes and the temporal reasoning used to retrieve evidence from them. \texttt{EgoScheme} converts local multimodal context into self-contained atomic memory indices, \texttt{EgoIndex} organizes complementary views without requiring global entity consolidation, and \texttt{EgoRetrv} combines soft temporal scoring with time-aware curation of retrieved evidence. Across the three-benchmark evaluation protocol, this formulation separates memory construction, evidence retrieval, and answer reasoning into inspectable stages. Future work should test the system under noisier perception, open-world identity resolution, and longer personal recordings.


\newpage


\clearpage
\appendix

\section{Baseline Failure Case Study}
\label{sec:appendix-baseline-failure}

\subsection{Failure in Memory Construction}
\label{sec:appendix-failure-resolution}

We ask whether existing memory representations produce self-contained and semantically complete memory indices. We analyze the memory indices constructed by two representative systems on EgoLife~\citep{yang2025egolife}: WorldMM~\citep{yeo2026worldmm}, which converts captions into OpenIE~\citep{angeli2015leveraging} \texttt{(subject, predicate, object)} triples, and LoCoMo~\citep{maharana2024evaluating}, which extracts speaker-attributed observation facts. We hypothesize that neither representation produces context-independent memories: WorldMM leaves referring expressions unresolved, while LoCoMo preserves utterances verbatim. To quantify this, we flag a WorldMM triple if its subject or object contains unresolved pronouns and a LoCoMo observation if it contains a quoted span, indicating that the original utterance is copied rather than resolved. Across 92,131 WorldMM triples and 254,997 LoCoMo observations, 10.1\% of WorldMM triples contain unresolved references and 17.4\% of LoCoMo observations contain verbatim quotes. We show examples in Tab.~\ref{tab:case-study-unresolved}. These results show that a substantial portion of existing memory indices are not self-contained, motivating a memory representation that explicitly resolves both coreference and ellipsis during memory construction.

\subsection{Failure in Time-Aware Retrieval}
\label{sec:appendix-failure-time-awareness}
We ask whether existing retrieval-based memory systems handle time-aware questions as effectively as time-unaware questions. To answer this, we construct a time-aware subset of EgoLifeQA (Appx.~\ref{sec:appendix-dataset-benchmark}) and evaluate two representative retrieval-based baselines, WorldMM~\citep{yeo2026worldmm} and VideoRAG~\citep{ren2026videorag}, using GPT-5.4 as the retrieval agent. Neither method explicitly models temporal intent during retrieval, instead relying primarily on semantic similarity. On the corresponding time-unaware questions, WorldMM and VideoRAG achieve 56.9\% and 56.7\% accuracy, respectively. However, their performance drops to 47.1\% (-9.8\%) and 43.4\% (-13.3\%) on time-aware questions. This substantial performance gap indicates that semantic retrieval alone is insufficient for long-horizon egocentric memory retrieval. We present representative failure cases in Fig.~\ref{tab:time-failures}.

\begin{table*}[t]
\centering
\small
\caption{Examples of unresolved memory produced by existing agentic memory baselines.}
\label{tab:case-study-unresolved}
\setlength{\tabcolsep}{5pt}
\renewcommand{\arraystretch}{1.15}

\begin{tabular}{p{0.15\textwidth}p{0.38\textwidth}p{0.39\textwidth}}
\toprule
\textbf{Baseline} & \textbf{Input Captions (excerpt)} & \textbf{Memory Entry (Triple / Observation)} \\
\midrule

\multirow{6}{*}{\textbf{WorldMM}}
& "Come on, everyone mark \textbf{it}." 
& [I $\mid$ says to Katrina, Alice, Tasha $\mid$ Come on, everyone mark \textbf{it}] \\

& "Keep \textbf{this} box." 
& [I $\mid$ say $\mid$ Keep \textbf{this} box] \\

& "\textbf{There} are a total of six computers." 
& [I $\mid$ states to Shure $\mid$ \textbf{there} are a total of six computers] \\

& "This thing, you can put \textbf{these} down." 
& [I $\mid$ say $\mid$ This thing, you can put \textbf{these}] \\

\midrule

\multirow{7}{*}{\textbf{LoCoMo}}
& Lucia: "Is it a little mouse?" \newline
Jake: "What is this?"
& [Jake] Jake asked "\textbf{What is this?}" \\

& Shure: "Great Northeast."
& [Shure] Shure said "\textbf{Great Northeast.}" \\

& Jake: "Or not, haha."
& [Jake] Jake said "\textbf{Or not, haha.}" \\

& Katrina: "Are all the packages out?" \newline
Jake: "Yes."
& [Jake] Jake replied "\textbf{Yes.}" \\

& Jake: "Or sit here."
& [Jake] Jake said "\textbf{Or sit here.}" \\

\bottomrule
\end{tabular}
\end{table*}

\begin{table}[t]
\centering
\small
\begin{tabular}{@{}p{0.20\linewidth}p{0.50\linewidth}p{0.2\linewidth}@{}}
\toprule
\textbf{Question ID} & \textbf{Question Statement} & \textbf{Time Cue} \\
\midrule
A1\_JAKE Q104   & When was the \emph{last time} cherries were mentioned? & ``last'' \\
A1\_JAKE Q1     & Who used the screwdriver \emph{first}? & ``first'' \\
A1\_JAKE Q90    & Where is the place I will \emph{never} go to? & ``never'' \\
A3\_TASHA Q903  & Who mic'd the chorus \emph{last night}? & ``last night'' \\
A3\_TASHA Q1091 & What did Lucia hand me \emph{last time}? & ``last'' \\
A3\_TASHA Q479  & Who do I \emph{usually} sit with when eating? & ``usually'' \\
A4\_LUCIA Q432  & Who mentioned AI \emph{last time}? & ``last'' \\
A4\_LUCIA Q256  & When did we \emph{first} mention juice? & ``first'' \\
A4\_LUCIA Q930  & Who bit the bullet \emph{yesterday morning}? & ``yesterday'' \\
A5\_KATRINA Q321  & What did I \emph{last} put on the shelf? & ``last'' \\
A5\_KATRINA Q111  & Who \emph{usually} participates when we select products? & ``usually'' \\
A5\_KATRINA Q1156 & How \emph{often} do I clean my room every day? & ``often'' \\
A6\_SHURE Q170  & When was the \emph{last time} I danced? & ``last'' \\
\bottomrule
\end{tabular}
\caption{Time-aware questions answered incorrectly by both WorldMM-GPT and VideoRAG. }
\label{tab:time-failures}
\end{table}

\section{Implementation Details}
\label{sec:appendix-implementation}

\subsection{Perception and Caption}
\label{sec:appendix-implementation-perception}
We primarily use the the human-annotated dense captions and speech transcripts as the source of memory in EgoLife~\citep{yang2025egolife} for fair baseline comparisons on agentic memory performance, following WorldMM~\citep{yeo2026worldmm}. For realistic evaluations, we also deploy VLMs as narrator to replace the dense captions as the source of video narrations as follows. 

\noindent\textbf{Person ID.}
We assume person identification is prior knowledge, since personal relationships are relatively stable and closed-set in real-world settings. This prior knowledge includes the mapping among a person's visual appearance, voice identity, and name. In the EgoLife dataset~\citep{yang2025egolife}, visual identity is difficult to obtain because \textit{faces are redacted in this dataset}, in contrast to EgoGPT and EgoButler~\citep{yang2025egolife}, which was developed by the dataset collectors with access to the raw videos. We therefore manually annotate each person's visual appearance, including clothing colors and styles, hairstyles, and other visual cues, across different recording scenes over the seven-day recordings (See example in Fig.~\ref{fig:person-id-example}). We generate identity files for each recording scene to map visual descriptions to person names. These identity files are used during VLM captioning to match people to their names. Empirically, we evaluate both proprietary VLM Gemini-3-Flash~\citep{team2023gemini} and open-weight VLM Gemma-4-31B~\citep{team2026gemma}. 

\begin{figure}[t]
\centering
\begin{promptbox}[Person ID Annotation Example: A1\_JAKE DAY1 11AM--5PM]
\begin{minted}[
    frame=none,
    framesep=3mm,
    linenos=false,
    tabsize=2,
    breaklines=true,
    breakanywhere=true,
    fontsize=\scriptsize,
    style=bw
]{text}
# Camera wearer (I)
I am Jake, the camera wearer.

# Other people who may appear
Katrina - pink long hair, white short-sleeve shirt, dark blue pants, yellow shoes
Alice - brown short hair, white long-sleeve t-shirt, grey skirt, grey shoes
Tasha - black long hair in a ponytail, black long-sleeve clothing, dark blue jeans
Lucia - long black hair, blue dress, blue shoes
Shure - male, short curly black hair, pink or orange short-sleeve shirt, brown pants, colorful shoes
\end{minted}
\end{promptbox}
\caption{Human expert annotated person ID example.}
\label{fig:person-id-example}
\end{figure}

\noindent\textbf{Video Caption.}
We first process 15-second, 1 FPS video clips independently from speech transcripts. Empirically, generating captions from joint video–speech inputs introduces hallucinations in objects, actions, and person identities, which we attribute to multimodal LLMs' tendency to prioritize textual over visual information~\citep{zheng2025mllms}. The resulting visual captions are then aligned with speech transcripts to construct multimodal captions, following WorldMM~\citep{yeo2026worldmm}.

\begin{table*}[t]
\centering
\small
\renewcommand{\arraystretch}{1.2}
\setlength{\tabcolsep}{5pt}

\begin{tabular}{p{0.18\textwidth} p{0.35\textwidth} p{0.4\textwidth}}
\toprule
\textbf{Component} & \textbf{Specification} & \textbf{Example} \\
\midrule

Category &
Extract \textbf{only} four event types:
(A) Hand--Object Interaction,
(B) Big Displacement,
(C) Utterance/Speech,
(D) Meaningful Gesture.
Discard all other events. &
"I fasten the shelf bracket with a screwdriver." \newline
"I walk into the kitchen." \newline
"Jake asks Tasha about the dinner menu." \newline
"Jake laughs out loudly." \\
\midrule

Atomicity &
Extract exactly one entry per distinct action.
Never combine actions connected by \emph{and}, \emph{while}, \emph{then}, or \emph{as}; split them into separate entries. &
\textbf{Good:}
"I'm working on computer to edit video." \newline
"Jake says weather will be bad tomorrow." \newline
\textbf{Bad:}
"I'm working on computer while listening to Jake talking about weather." \\
\midrule

Action Format &
Each entry should contain 8--20 words in the form \emph{subject + verb + object}. Use "I" for the camera wearer and a real name for everyone else. Preserve named objects and places verbatim. &
"Shure hands me the screwdriver at desk." \newline
"I message Lucia on my phone." \newline
"I fasten the shelf bracket with a screwdriver." \\
\midrule

Entity Resolution &
Resolve every pronoun and generic coreference to an explicit person, object, or place. Discard an entry only if the referent is truly unresolvable. &
\textbf{Good:} "Jake picks up the bottle on table." \newline
\textbf{Bad:} "He picks it up." \\
\midrule

Speech Normalization &
Convert dialogue into a speech act with a complete, self-contained topic expressed using a WHAT or THAT clause. &
\textbf{Good:}
"Jake tells Tasha that dinner is ready." \newline
"I ask Alice about the shelf height." \newline
\textbf{Bad:}
"Jake says OK." "I ask Alice about it." 
\\
\midrule

Motion Filtering &
Discard micro-gestures and undirected motion. Keep movement only when it reaches a named destination. &
\textbf{Good:}
"I walk into the kitchen." \newline
\textbf{Bad:} "I walk forward." \\
\midrule

Output &
Return JSON only. &
\texttt{\{"actions": ["entry1", "entry2", ...]\}} \\

\bottomrule
\end{tabular}

\caption{Components of the action and utterance extraction prompt.}
\label{tab:prompt-action-utterance}
\end{table*}

\subsection{index Memory Construction}
\label{sec:appendix-implementation-construction}

\noindent\textbf{Action \& Utterance.}
Given the generated 5-minute context window, we resolve coreference and ellipsis following \texttt{EgoScheme} to construct action and utterance atomic memory indices from multimodal captions. Specifically, the context-resolution LLM receives the current caption together with its preceding 5-minute context, enabling it to rewrite each memory index into a self-contained and semantically complete representation. The LLM then extracts structured action and utterance indices according to the prompt in Fig.~\ref{fig:action-utterance-extraction}. Action indices capture hand-object interactions, large movements, and meaningful gestures. Utterance indices are normalized into explicit speech acts or statements with resolved speakers, referents, and topics. Compared with the original captions, action indices become more specific by grounding actions to explicit objects and locations. Utterance indices become context-independent by resolving pronouns and ellipsis. Finally, the resulting indices are embedded using Qwen3-Embedding-4B~\citep{yang2025qwen3} and indexed with FAISS~\citep{douze2025faiss}.

\noindent\textbf{Activity \& Conversation.} 
Following WorldMM, we prompt an LLM to extract coarse-grained, goal-oriented activities from each 30-minute multimodal context, as shown in Fig.~\ref{fig:activity-extraction}. The LLM may generate one or more activity indices, each associated with a specific time range. We observe that the generated activities are often coupled, using conjunctions such as ``and'' or ``while'' to combine multiple unrelated activities into a single index. Such coupled indices are difficult to retrieve semantically, while naively shortening them often removes important details. We therefore introduce a disentangling harness that separates coupled activities while preserving the original semantics. To reconcile activities across adjacent 30-minute segments, we use Qwen3-Embedding-4B~\citep{yang2025qwen3} to identify temporally continuous activities with a cosine similarity above 0.9, followed by an additional LLM call to merge them. The resulting activity indices are embedded and indexed using the same backend.
Conversation indices are constructed in the same manner. We prompt the LLM to extract coarse-grained, goal-oriented conversations from each 30-minute multimodal context, as shown in Fig.~\ref{fig:conversation-extraction}. The extracted conversations are disentangled, merged across segment boundaries, and embedded into the conversation memory using the same pipeline.

\subsection{Drafting Agent}
\label{sec:appendix-implementation-drafting-agent}

\noindent\textbf{Index Pool.}
The \textit{index pool} stores all atomic memory indices retrieved from the database across retrieval rounds. Whenever new indices are retrieved, they are added to the pool, deduplicated by content, and sorted chronologically by timestamp. At the beginning of each retrieval round, the drafting agent receives the entire index pool together with the newly retrieved indices, allowing it to progressively refine subsequent retrieval queries.

\noindent\textbf{Tool Calling.}
We use the agentic tool calling capabilities of the latest LLMs for the drafting agent. We employ persistent thinking mode in Qwen models and function calling in GPT models for multi-round Chain-of-Thought (CoT) reasoning and atomic memory index drafting. In each round, the drafting agent decides one of five tool calls: \verb|search_action_utterance|, \verb|search_activity|, \verb|search_speech|, \verb|curate_evidence|, or \verb|answer|. We present the detailed tool call schema in Fig.~\ref{fig:tool-call-scheme}.

\verb|search_action_utterance| takes one required \verb|query| for short-term fine-grained action or utterance vector database search, which should include specific objects, persons, and/or utterances. \verb|search_activity| takes one required \verb|query| for long-term coarse-grained activity vector database search. \verb|search_conversation| takes one required \verb|query| for long-term coarse-grained conversation vector database search, which should include specific conversational topics. All three tools take an optional \verb|time_query|, which indicates the most likely occurrence time of the searched events. It is captured by the database harness and translated into the standard timestamp format \verb|DAYX HHMMSSFF|. This enables the agent to perform time-aware retrieval and query refinement for adaptive searching. Each time the database returns atomic memory index, the harness stores the index in a working memory \textit{index pool}. Meanwhile, new retrieved atomic memory indices and the existing index pool are added to the agent context for the next round of CoT reasoning.

\verb|curate_evidence| is a drafting-agent curation tool used to keep the \textit{index pool} clean and organized, rather than the later curation agent. The agent can call this tool with a list of the retrieved memory indices to decide which memory indices to keep. The harness captures the list of indices to keep and removes the rest from the \textit{index pool}. Empirically, we observe that strong agents such as GPT-5.4 and Sonnet-4.6~\citep{anthropic2026sonnet46} are good at curating indices, while weaker agents such as Qwen3.6-27B frequently make mistakes and become biased toward certain indices. We therefore only enable this tool call for strong agents. \verb|answer| is the tool used to terminate the drafting agent CoT. It collects all indices in the \textit{index pool} and sends it to the curation agent for final index curation before question answering.

\noindent\textbf{Harness.}
\verb|Query| and \verb|curate_evidence| are automatically captured by the harness through string and array-list parsing. \verb|time_query| is parsed into a canonical timestamp format, \verb|DAYX HHMMSSFF|, which can represent either a timestamp or a time range. Since the dataset recordings can be intermittent, we concatenate the recordings into a single uniform timeline. Following Eq.~\ref{eq:time-decay}, we map the timestamps of each candidate memory index in the vector database to a time-relevance score. We further compute the semantic similarity score by embedding the \verb|query| using Qwen3-Embedding-4B model and search it in the pre-embedded FAISS vector database~\citep{douze2025faiss}. The resulting semantic similarity score and time-relevance score are multiplied as the final ranking score. 

\subsection{Sampling Agent}
\label{sec:appendix-implementation-sampling-agent}
The sampling agent is implemented as a single LLM call that curates the atomic memory indices in the \textit{index pool}. Its goal is to keep the reasoning context of the final response agent concise and relevant. We prompt the agent to explicitly reason over temporal qualifiers in the question, such as ``usually'', ``habit'', ``last'', and ``this morning'' (see prompt in Fig.~\ref{fig:curation-agent-prompt}). Based on the inferred temporal intent, the agent selects memory indices that best match the question. For example, habitual questions favor diverse memories across time, whereas recency questions prioritize the most recent indices. Conversely, indices outside the relevant temporal scope are discarded. In practice, the sampling agent typically returns fewer than 15 atomic memory indices for the response agent.

\subsection{Reasoning \& Response}
\label{sec:appendix-implementation-response}

The response agent receives the curated atomic memory indices and maps their timestamps back to the corresponding multimodal captions. Given the question, answer options, query timestamp, and retrieved captions, it performs multimodal reasoning and generates the final answer.

\subsection{Models \& Configurations}
\noindent\textbf{Memory Perception \& Construction.}
We evaluate three experimental setups for memory perception and construction. (1) Dense captions: GPT-5.4 (without thinking) is used for all memory construction calls over the human-annotated dense captions. (2) VLM narration: Gemini-3-Flash (without thinking) generates video narrations, followed by GPT-5.4 (without thinking) for memory construction. (3) Open-source pipeline: Gemma-4-31B is used for both video captioning and memory construction.

\noindent\textbf{Retrieval \& Response.}
We primarily evaluate \texttt{EgoCITE} on two agentic LLMs, GPT-5.4 and Qwen3.6-27B-FP8 with both tool calling mode. 
GPT-5.4 enables both function calling (tool calling) and thinking mode at medium effort with maximum 4096 output tokens for drafting agent and sampling agent. The response agent enables thinking mode at medium with maximum 4096 output tokens without tool calling. 
Qwen3.6-27B-FP8 enables both persistent thinking (tool calling) and thinking mode with maximum 4096 output tokens for drafting agent and sampling agent. The response agent enables thinking mode with maximum 4096 output tokens without tool calling. 
Both models we use the top-$k$ values for retrieval with action and utterance memory at $k=20$, activity memory at $k=10$, and conversation memory $k=10$. We set the drafting agent tool-calling round number to 5 with at most 15 retrieved atomic memory indices to align with WorldMM~\citep{yeo2026worldmm}.


\section{Experiment Details}
\label{sec:appendix-evaluation}

\subsection{Datasets \& Benchmarks}
\label{sec:appendix-dataset-benchmark}

\textbf{EgoLife}\footnote{https://huggingface.co/datasets/lmms-lab/EgoLife} is a socially intensive daily lifelogging dataset collected from six participants, comprising 7 days $\times$ 8 hours of continuous multimodal recordings per person, including egocentric RGB video and audio~\citep{yang2025egolife}. EgoLife provides human-annotated dense captions describing both first-person and third-person actions, as well as speech transcripts. Since the dataset is primarily in Chinese, we translate all dense captions and transcripts into English using Qwen3.6-27B-FP8~\citep{qwen3.6}. Following the practice of WorldMM~\citep{yeo2026worldmm}, we further align the translated captions and transcripts into 30-second windows to construct the multimodal captions used as the primary inputs to the memory systems. This dataset's content is used in three benchmarks: EgoLifeQA~\citep{yang2025egolife}, EgoR1-Bench~\citep{tian2025ego}, and EgoMem~\citep{zheng2026lifedialbench}. 

\noindent\textbf{EgoLifeQA}\footnote{https://huggingface.co/datasets/Ego-R1/Ego-R1-Data} contains 2,905 manually annotated multiple-choice questions from EgoLife~\citep{yang2025egolife} with five categories: Entity Log, Event Recall, Habit Insight, Relation Map, and Task Master. Each question includes one question statement, four answer choices, the question time, the ground-truth answer, and a target timestamp indicating the timestamps needed to answer the question. We use the question statement, answer choices, and question time as the question context for retrieval, curation, and response agent. We use the ground-truth answer and target timestamp as evaluation metrics. 

\noindent\textbf{EgoLifeQA Time-Aware Questions.} Additionally, we split the EgoLifeQA benchmark into two subsets: \textit{time-aware questions} and \textit{time-unaware questions}. Time-aware questions contain keywords that clearly indicate temporal cues, including "usually", "often", "never", "morning", "afternoon", "DAY X", and "last", etc. (see Tab.~\ref{tab:time-relevant-questions}). This subset contains 2,223 questions, covering 76\% of the EgoLifeQA benchmark.

\begin{table}[t]
  \centering
  \caption{Keywords used to classify temporal-aware questions in EgoLifeQA.}
  \label{tab:time-keywords}
  \begin{tabular}{@{}ll@{}}
    \toprule
    Category & Keywords \\
    \midrule
    Time of day       & morning, noon, afternoon, evening, night, midnight, \\
                      & dawn, dusk, tonight \\
    Calendar / days   & today, yesterday, tomorrow, day, days, date, week, \\
                      & weekend \\
    Frequency / habit & usually, often, always, never, sometimes, frequently, \\
                      & rarely, routinely, regularly, habit \\
    Ordering / recency& first, last, latest, earlier, earliest, before, previous, \\
                      & previously, recent, recently, ago, then, initially, \\
                      & originally, prior, final, finally \\
    Generic temporal  & when, again, already, just, past, moment, hour, \\
                      & minute, o'clock, order, sequence \\
    \bottomrule
  \end{tabular}
  \label{tab:time-relevant-questions}
\end{table}

\noindent\textbf{EgoMem}\footnote{https://github.com/RayNeo-AI-2025/LifeDialBench} contains 939 multiple-choice questions from LifeDialBench~\citep{zheng2026lifedialbench} based on contents of EgoLife~\citep{yang2025egolife}, including four categories: Single Event, Multiple Event, Event Detail, and Time Query. Each question includes a question statement, four answer choices, the question time, the ground-truth answer, and a target timestamp indicating the timestamps needed to answer the question. We use the question statement, answer choices, and question time as the question context for retrieval, curation, and response generation. The ground-truth answer and index timestamp are used for evaluation. Since the questions are written in third person, where camera wearers are referred to by name rather than as "I", we use GPT-5.4~\citep{openai2026gpt54thinking} to normalize the questions and choices from third person to first person. To further align with EgoLifeQA~\citep{yang2025egolife}, we normalize the date and time into the same format, e.g., "DAY1" for dates and "12000000" for timestamps.

\subsection{Metrics}
\label{sec:appendix-metrics}

\textbf{Accuracy} evaluates whether the response agent selects the correct choice among the candidate answers compared with the ground-truth answer. We prompt the response agent to generate a JSON output containing its prediction. Since LLMs may produce invalid JSON outputs, we implement an additional answer extraction step using an LLM call to extract the predicted choice.

\noindent\textbf{Hits and NearHits} evaluate \textit{multi-round} retrieval performance by measuring whether the retrieved memory indices are temporally aligned with, or close to, the target multimodal captions needed to answer the question. \textit{We use these metrics to evaluate multi-round retrieval agent performance.} We define \textit{Hits} as whether the retrieved memory indices, temporally overlap with the target time. We define \textit{NearHits} as whether the retrieved memory indices temporally overlap with a $\pm 5$ minute window around the target time. This metric is motivated by the observation that answers can often still be inferred from nearby context even when the exact target time is missed. However, \textit{Hits} and \textit{NearHits} do not necessarily guarantee the correctness of the final answer.

\noindent\textbf{Normalized Cost} measures retrieval efficiency using the normalized API token cost of GPT-5.4~\citep{openai2026pricing}. According to the GPT-5.4 pricing, short-context models incur a normalized cost of $1\times$ and $6\times$ per input and output token, respectively, while long-context models incur $2\times$ and $9\times$. \texttt{EgoCITE-GPT} and other GPT-based agentic memory baselines use short-context pricing, whereas the long-context GPT agent uses long-context pricing. The normalized cost is computed as the weighted sum of input and output token costs.

\subsection{Baselines}

\textbf{Long-context LLM Agents.} State-of-the-art LLMs, such as GPT-5.4~\citep{openai2026gpt54thinking} and Gemini-3.1-Pro~\citep{team2023gemini}, provide built-in agentic capabilities through tool use (e.g., keyword search and Python execution). We therefore treat them as long-context LLM agent baselines. Since the 56-hour video recording exceeds the maximum context window of commonly used VLMs, we use JSON files containing timestamps and multimodal captions as the inputs to LLMs with 1M-token context windows. We set the thinking effort of GPT-5.4 to medium and Gemini-3.1-Pro to medium. The LLMs are prompted to gather index from the input JSON documents and answer the questions. Each JSON document consumes 0.7--0.9M input tokens. To efficiently evaluate the questions, we employ a batch prompting strategy~\citep{cheng2023batch}, grouping 50 questions at a time. Questions without valid answers are re-processed. For latency and cost evaluation, we randomly sample 100 questions across the three benchmarks and prompt each question independently to obtain unbiased per-question efficiency measurements.

\noindent\textbf{EgoRAG}\footnote{https://github.com/EvolvingLMMs-Lab/EgoLife} uses multi-granularity summaries as the indices of the vector database~\citep{yang2025egolife}. During memory construction, we use GPT-5.4 to generate summaries from multimodal captions at multiple granularities and store them in the vector database. During retrieval, we use GPT-5.4 to generate vector database queries, and use GPT-5.4 with medium thinking effort to answer the questions. We primarily follow the original codebase, except for replacing the backend LLMs.

\noindent\textbf{A-MEM}\footnote{https://github.com/agiresearch/a-mem} uses keyword tags as the memory index~\citep{xu2026mem}. During memory construction, we use GPT-5.4 to extract keywords and tag the multimodal captions accordingly. During retrieval, we use GPT-5.4 to generate queries, and use GPT-5.4 with medium thinking effort to answer the questions. We retrieve the top-$k=15$ captions. We primarily follow the original codebase, except for replacing the backend LLMs and providing keyword extraction, retrieval, and response prompts from EgoRAG.

\noindent\textbf{VideoRAG}\footnote{https://github.com/HKUDS/VideoRAG} employs both knowledge graph entity and visual vector for memory indices. At retrieval stage, it utilizes both semantic similarity comparison and graph walk to retrieve relevant captions and video clips. We use the multimodal captions for entity knowledge graph construction and videos for visual memory. We primarily follow the code base except replacing GPT-4o model in entity extraction and final answering stage into GPT-5.4 model and GPT-4o model in retrieval flows into GPT-5.4 model. We reuse the visual embeddings from WorldMM built by VLM2Vec for visual memory. 

\noindent\textbf{WorldMM}\footnote{https://github.com/wgcyeo/WorldMM} employs three types of memory, including episodic, semantic, and visual memory, together with multi-round adaptive retrieval~\citep{yeo2026worldmm}. Episodic and semantic memory are stored in knowledge graphs and visual memory is stored in vector database. We primarily follow the pipeline of the original codebase. We use GPT-5.4 to construct the three memory types from multimodal captions instead of GPT-5-mini. Similarly, we use GPT-5.4 instead of GPT-5 as the retriever, and GPT-5.4 with medium reasoning effort instead of GPT-5 for final question answering. The visual memory is built by visual embedding model VLM2Vec. We set the maximum number of retrieval rounds to 5 and maximum number of memory indices retrieved as 15, following default implementation of WorldMM.

\subsection{Hardware}
All small embedding models, including the Qwen3-Embedding-4B text embedding model and the VLM2Vec visual embedding model, are processed locally on a machine equipped with an RTX 4090 GPU, an Intel i5-13600KF CPU, and 128GB RAM. The Qwen3.6-27B-FP8 and Gemma-4-31B model are served on a cloud server equipped with 4$\times$ RTX 6000 Pro GPUs using vLLM~\citep{kwon2023efficient}.


\section{Additional Results}
\label{sec:appendix-additional-results}

\begin{table}[t]
\centering
\caption{Component ablation study.}
\label{tab:component-ablation}
\setlength{\tabcolsep}{3pt}
\begin{tabular}{lcccc}
\toprule
\textbf{Metric} &
\textbf{Caption RAG} &
\textbf{+EgoIndex} &
\textbf{+EgoScheme} &
\textbf{+EgoRetrv} \\
\midrule
Accuracy (\%) & 61.3 & 69.7 & 71.3 & 75.3 \\
Hit Rate (\%) & 40.7 & 54.3 & 59.3 & 62.7 \\
\bottomrule
\end{tabular}
\end{table}

\begin{table}[t]
\centering
\caption{Time-decay factor $\lambda$ ablation on \texttt{EgoCITE-GPT}.}
\label{tab:decay-lambda-ablation}
\setlength{\tabcolsep}{3pt}
\begin{tabular}{lcccc}
\toprule
\textbf{Metric} & $\lambda=0.97$ & $\lambda=0.98$ & $\lambda=0.99$ & $\lambda=0.995$ \\
\midrule
Accuracy (\%) & 72.0 & 76.0 & \textbf{81.0} & 76.0 \\
Hit Rate (\%) & 56.0 & 55.0 & \textbf{67.0} & 54.0 \\
\bottomrule
\end{tabular}
\end{table}

\noindent\textbf{NearHits.}
We further evaluate the retrieval agent using the NearHits metric (Tab.~\ref{tab:near-hit}), where retrieval is considered successful if any of the at most 15 retrieved memory indices fall within a $\pm$5-minute window of the ground-truth evidence. \texttt{EgoCITE-GPT} achieves the highest NearHits rates of 69.4\%, 90.8\%, and 70.3\% on EgoLifeQA, EgoMem, and EgoR1, respectively, improving over the strongest baseline by up to 28.7\%, 14.8\%, and 24.3\%. Compared with the Hit Rate results, the consistently higher NearHits rates indicate that \texttt{EgoCITE} frequently localizes evidence to the correct temporal neighborhood even when the exact target timestamp is not retrieved.

\begin{table}[t]
\centering
\caption{Retrieval NearHits rate (\%). NearHits indicate the retrieved memory indices fall into the $\pm$5-minute margin of ground truth evidence.}
\label{tab:near-hit}
\setlength{\tabcolsep}{3pt}
\renewcommand{\arraystretch}{1.05}
\begin{tabular}{lccc}
\toprule
\textbf{Method} & \textbf{EgoLifeQA} & \textbf{EgoMem} & \textbf{EgoR1} \\
\midrule
EgoRAG~\citep{yang2025egolife}      & 14.5 & 44.9 & 13.5 \\
A-Mem~\citep{xu2026mem}             & 33.2 & 46.6 & 36.3 \\
VideoRAG~\citep{ren2026videorag}    & 6.2  & 19.7 & 9.7  \\
WorldMM-Qwen~\citep{yeo2026worldmm} & 39.5 & \underline{76.0} & 46.0 \\
WorldMM-GPT~\citep{yeo2026worldmm}  & 40.7 & 73.5 & 45.3 \\
\midrule
\textbf{\texttt{EgoCITE-Qwen} (Ours)} &
  \underline{63.6} &
  \textit{75.6} &
  \underline{68.3} \\
\textbf{\texttt{EgoCITE-GPT} (Ours)} &
  \textbf{69.4}\textcolor{green!60!black}{\small(+28.7)} &
  \textbf{90.8}\textcolor{green!60!black}{\small(+14.8)} &
  \textbf{70.3}\textcolor{green!60!black}{\small(+24.3)} \\
\bottomrule
\end{tabular}
\end{table}

\noindent\textbf{Retrieval Efficiency and Cost.}
We report the per-question input tokens, output tokens, latency, and API cost in Tab.~\ref{tab:tradeoff-breakdown}. \texttt{EgoCITE-GPT} offers a tunable accuracy-efficiency trade-off. The 1-round variant achieves 58.7\% accuracy using only 9.3k input tokens, \$0.034 per query, and 15.2s latency. Increasing retrieval depth to five rounds improves accuracy by 5.1\% at the cost of 3.4$\times$ more input tokens, 2.1$\times$ higher latency, and 3.2$\times$ higher cost. Compared with GPT-5.4, the default 5-round configuration achieves higher accuracy while using 24.7$\times$ fewer input tokens and 36$\times$ lower cost with comparable latency. It also requires 2.3$\times$ fewer input tokens, 2$\times$ lower cost, and 14.8s lower latency than WorldMM-GPT while achieving substantially higher accuracy. Overall, \texttt{EgoCITE} achieves an effective balance between accuracy, latency, and computational cost.

\begin{table}[t]
\centering
\caption{Per-question cost breakdown for the latency--accuracy tradeoff. Cost is computed based on GPT-5.4 API pricing~\citep{openai2026pricing}.}
\label{tab:tradeoff-breakdown}
\setlength{\tabcolsep}{2pt}
\begin{tabular}{lrrrr}
\toprule
\textbf{Method} &
\textbf{Input tok.} &
\textbf{Output tok.} &
\textbf{Latency (s)} &
\textbf{Cost per Q (\$)} \\
\midrule
\textbf{\texttt{EgoCITE-GPT} (1 Round)} & 9,288   & 743   & 15.2 & 0.034 \\
\textbf{\texttt{EgoCITE-GPT} (5 Round)} & 31,670  & 2,073 & 32.0 & 0.110 \\
GPT-5.4 agent~\citep{openai2026gpt54thinking}
                               & 783,123 & 2,826 & 32.5 & 3.979 \\
VideoRAG~\citep{ren2026videorag}
                               & 21,100  & 3,162 & 26.4 & 0.100 \\
WorldMM-GPT~\citep{yeo2026worldmm}
                               & 71,801  & 2,327 & 46.8 & 0.214 \\
A-Mem~\citep{xu2026mem}
                               & 1,598   & 805   & 9.6  & 0.016 \\
EgoRAG~\citep{yang2025egolife}
                               & 15,914  & 525   & 10.6 & 0.048 \\
\bottomrule
\end{tabular}
\end{table}

\noindent\textbf{Memory Granularity Ablation.}
We ablate the multi-view design of \texttt{EgoIndex} in \texttt{EgoCITE-GPT} by varying memory granularity on EgoR1-Bench. Fine-grained memory (actions and utterances) achieves 69.7\% accuracy and 55.7\% hit rate, while coarse-grained memory (activities and conversations) reaches 73.3\% and 59.3\%, respectively. Combining all four views further improves accuracy to 75.3\% and hit rate to 62.7\%. These results demonstrate that fine-grained and coarse-grained memories provide complementary information for long-horizon retrieval and QA.

\noindent\textbf{Video Caption Results.}
We replace the default human-annotated dense-caption memory source with video narrations generated by Gemini-3-Flash and open-weight Gemma-4-31B. The Gemini pipeline uses GPT-5.4 for memory construction, consistent with the dense-caption setting, while the Gemma pipeline uses Gemma-4-31B throughout memory construction to evaluate a fully open-weight setting. All variants use Qwen3.6-27B-FP8 as the retrieval agent and are evaluated on EgoLifeQA. As shown in Tab.~\ref{tab:video-caption}, replacing human-annotated dense captions with VLM-generated narrations results in only a modest performance drop. Gemini captions achieve 60.3\% accuracy and 40.7\% hit rate, only 1.2\% and 2.9\% lower than dense captions, respectively, while the fully open-weight Gemma pipeline achieves 57.1\% accuracy and 36.3\% hit rate. These results demonstrate that \texttt{EgoCITE} generalizes well to automatically generated video narrations, enabling practical deployment without relying on human-annotated dense captions.

\begin{table}[t]
\centering
\caption{Video captioning variants comparison to dense caption baseline on EgoLifeQA.}
\label{tab:video-caption}
\setlength{\tabcolsep}{2pt}
\begin{tabular}{lccc}
\toprule
\textbf{Metric} & \textbf{Human-annotated} & \textbf{Proprietary VLM} & \textbf{Open-source VLM} \\
\midrule
Video narrator &
Dense Caption &
Gemini-3-Flash &
Gemma-4-31B \\
Memory constructor &
GPT-5.4 &
GPT-5.4 &
Gemma-4-31B \\
\midrule
Hit rate (\%)         & 43.6 & 40.7 & 36.3 \\
Average accuracy (\%) & 61.5 & 60.3 & 57.1 \\
\midrule
EntityLog    & 62.7 & 61.4 & 56.0 \\
EventRecall  & 59.5 & 57.6 & 53.8 \\
HabitInsight & 58.8 & 59.8 & 59.1 \\
RelationMap  & 63.6 & 60.5 & 59.9 \\
TaskMaster   & 67.1 & 70.1 & 68.2 \\
\bottomrule
\end{tabular}
\end{table}


\section{Future Work and Discussion}

Although \texttt{EgoCITE} demonstrates that memory representation is critical for long-horizon egocentric retrieval, several challenges remain. 

\textbf{Multimodal Perception.} Memory quality is still bounded by multimodal perception, particularly person identification under privacy-preserving face-redaction settings. An individual's appearance (e.g., clothing and hairstyle) may change substantially over long time horizons, making reliable visual identification difficult for both CV models and VLMs. Additionally, our system assumes accurate ASR transcripts with resolved speaker identities. In realistic settings, however, ASR errors, overlapping speech, and speaker diarization failures may further degrade memory construction and retrieval. Improving robust multimodal perception remains an important direction for future egocentric memory systems.

\textbf{Cross-Modality Index Construction.} We resolve coreference and ellipsis over text-based multimodal captions using LLMs, treating captions as an intermediate representation of the underlying multimodal signals. This abstraction inevitably loses cross-modal cues that are useful for reference resolution, such as gaze, pointing gestures, speaker localization, and object interactions. Future work could instead perform coreference and ellipsis resolution directly over synchronized multimodal signals, allowing speech, vision, and actions to jointly ground entities, events, and omitted references. Such cross-modal memory construction would produce richer memory indices and reduce the need to densely caption and index long-horizon videos, improving both efficiency and retrieval quality.

\textbf{Memory Structure.} We adopt a simple vector database to isolate the effect of memory representation in this work. We believe that the quality of the underlying memory indices is a prerequisite for any memory system, regardless of whether the indices are stored in a vector database, knowledge graph, or other structured memory. Richer memory structures may offer additional benefits, such as relational reasoning, hierarchical organization, or more efficient graph traversal. Future work could therefore integrate the proposed memory indexing scheme with sophisticated memory structures, enabling them to reason over cleaner, self-contained, and contextually complete memory indices.

\textbf{Reasoning and Guessing.} Hit and near-hit rates measure retrieval alignment but do not fully determine QA correctness. On EgoLifeQA, \texttt{EgoCITE-GPT} and \texttt{EgoCITE-Qwen} retrieve the target evidence but answer incorrectly on 11.3\% and 9.1\% of questions, while 12.6\% and 15.3\% are answered correctly despite missing the annotated evidence. These mismatches highlight two directions for future work: (1) improving agent reasoning over correctly retrieved memories to reduce reasoning failures, and (2) improving benchmark questions and ground-truth evidence annotations, which may contain annotation noise or non-exclusive evidence that allows an answer to be inferred or guessed from memories outside the annotated target.

\section{Examples and Comparisons}
\label{sec:appendix-examples}

\begin{figure*}[h!]
    \centering
    \includegraphics[width=\textwidth]{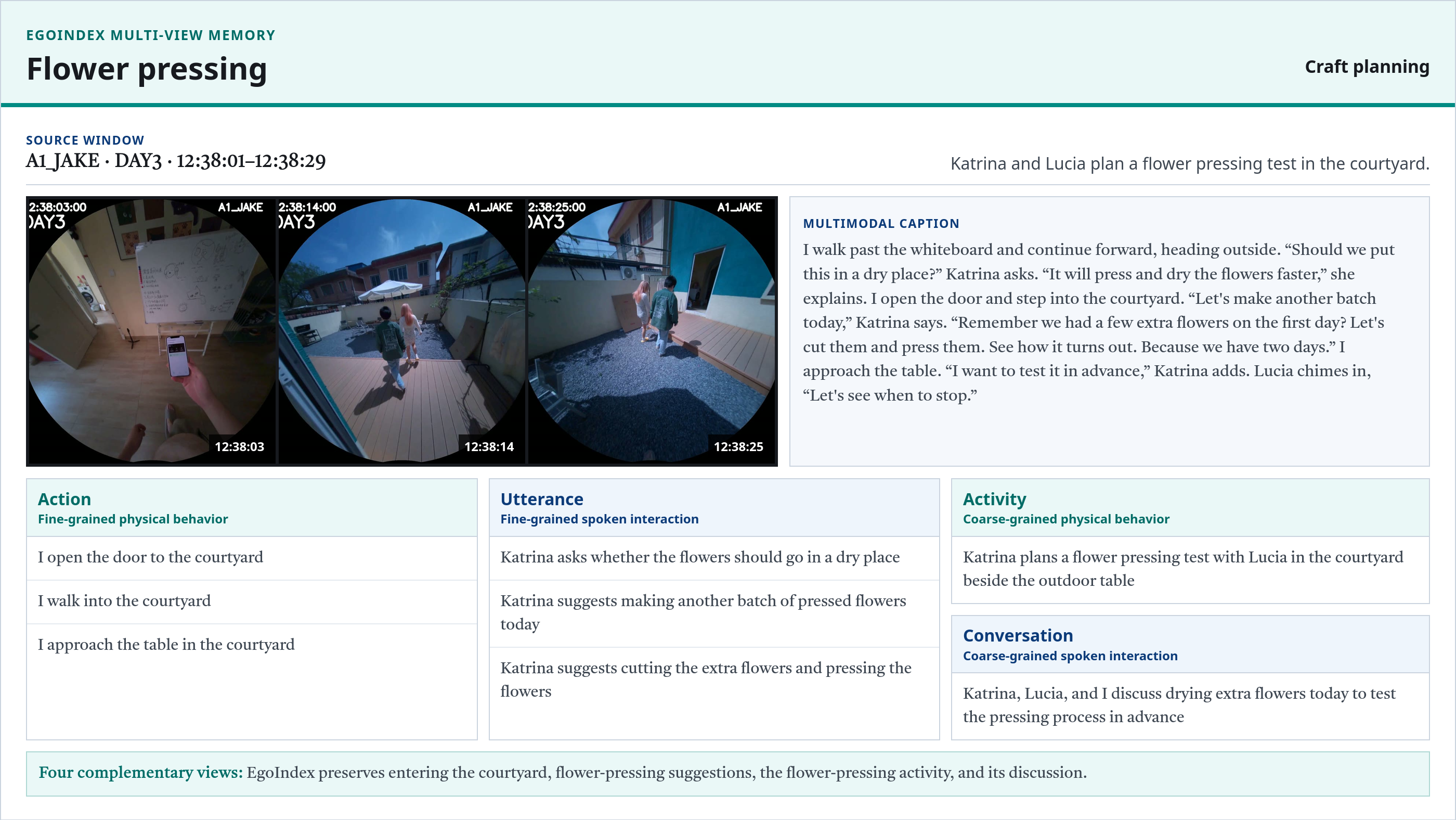}
    \caption{EgoIndex multi-view memory example for flower pressing. From one caption window, EgoIndex constructs fine-grained action and utterance indices alongside coarse-grained activity and conversation indices.}
    \label{fig:egoindex-flower-pressing}
\end{figure*}

\begin{figure*}[h!]
    \centering
    \includegraphics[width=\textwidth]{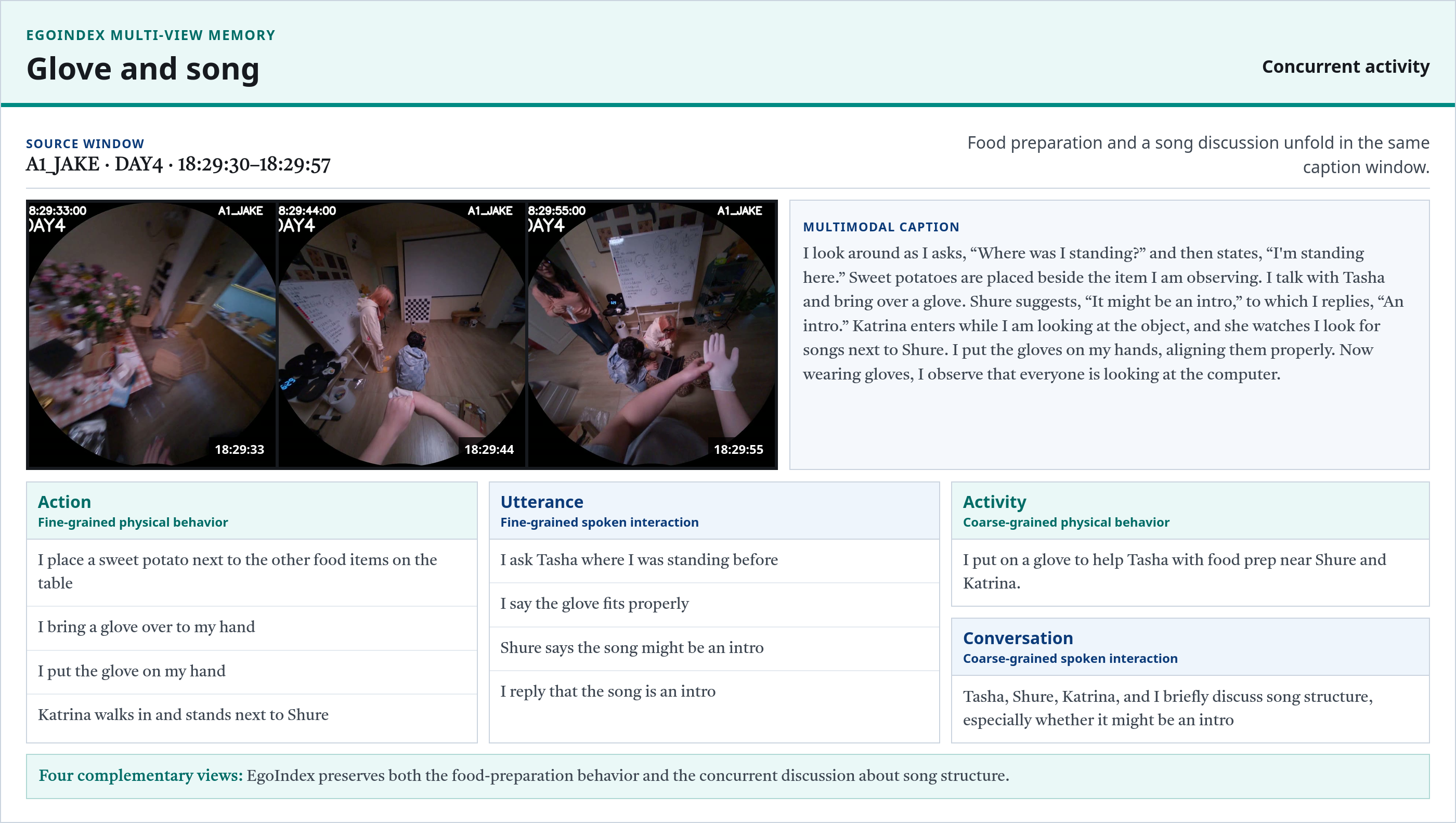}
    \caption{EgoIndex multi-view memory example for concurrent activity. EgoIndex separately preserves food-preparation actions, utterances about the song, the broader food-preparation activity, and the conversation about song structure.}
    \label{fig:egoindex-glove-and-song}
\end{figure*}

\begin{figure*}[h!]
    \centering
    \includegraphics[width=\textwidth]{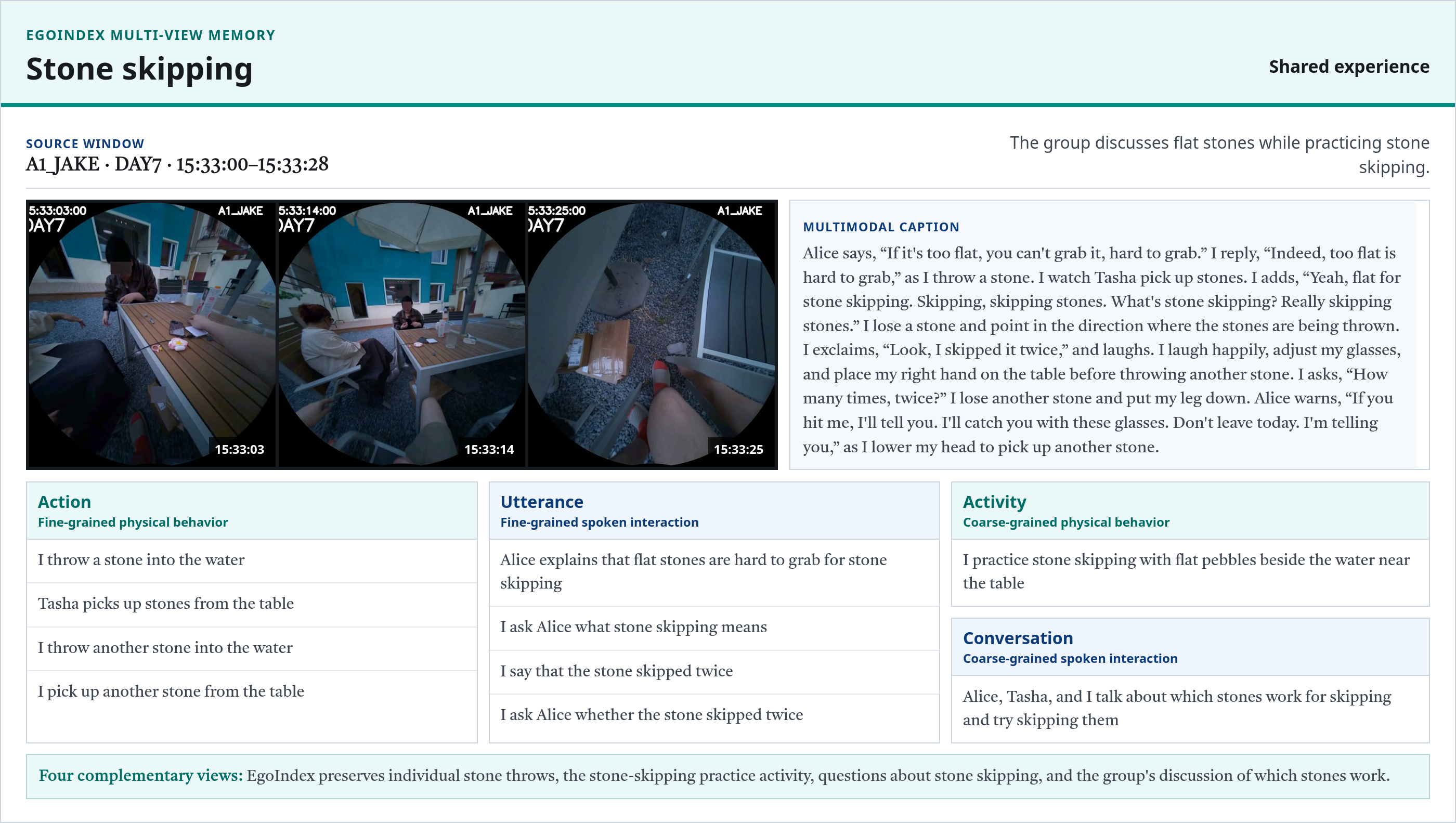}
    \caption{EgoIndex multi-view memory example for stone skipping. EgoIndex preserves individual stone-related actions and utterances together with the broader stone-skipping activity and conversation topic.}
    \label{fig:egoindex-stone-skipping}
\end{figure*}

\begin{figure*}[h!]
    \centering
    \includegraphics[width=\textwidth]{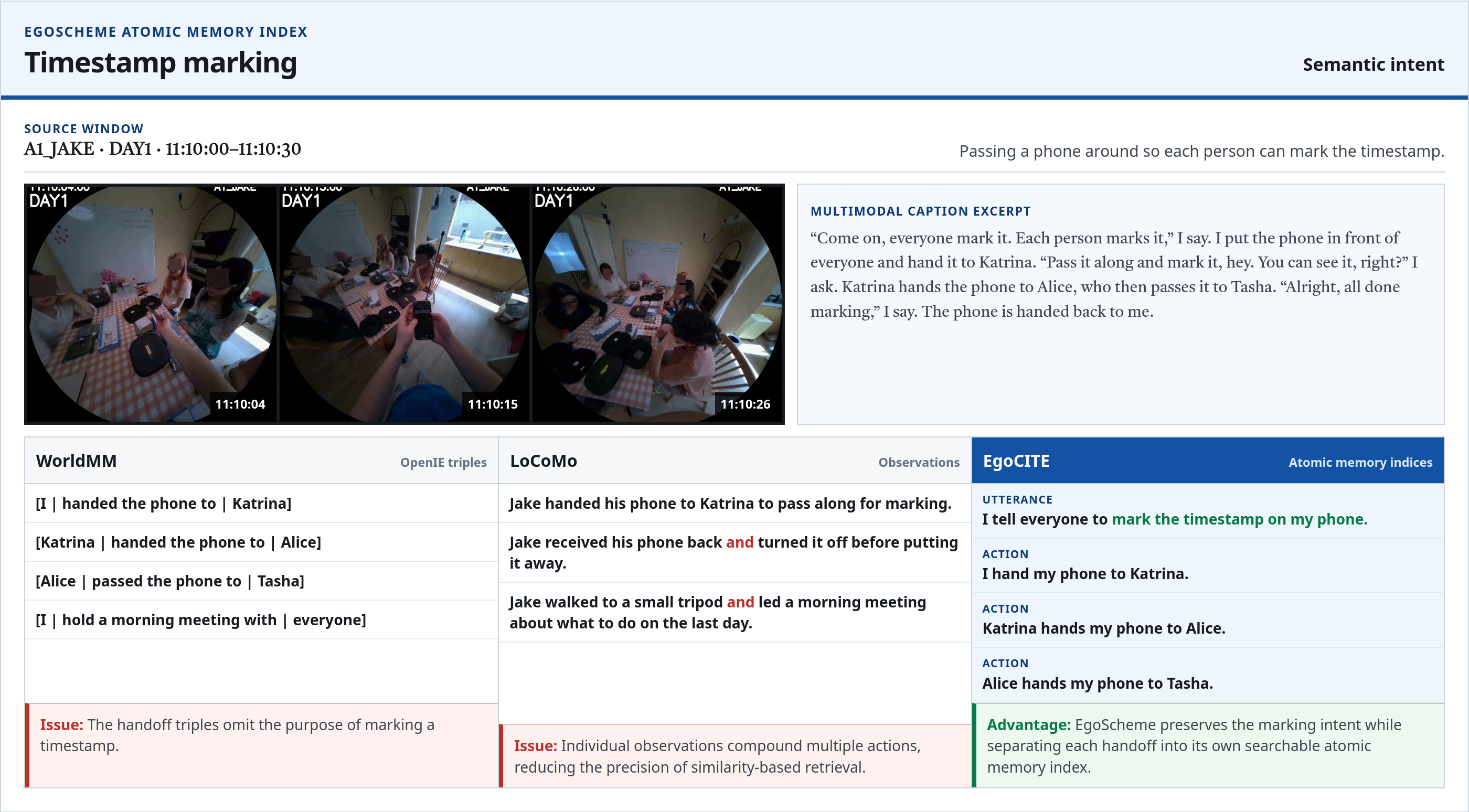}
    \caption{EgoScheme example for timestamp marking. Compared with WorldMM's context-poor triples and LoCoMo's compounded observations, EgoScheme preserves the marking intent and separates the phone handoffs into atomic memory indices.}
    \label{fig:egoscheme-timestamp-marking}
\end{figure*}

\begin{figure*}[h!]
    \centering
    \includegraphics[width=\textwidth]{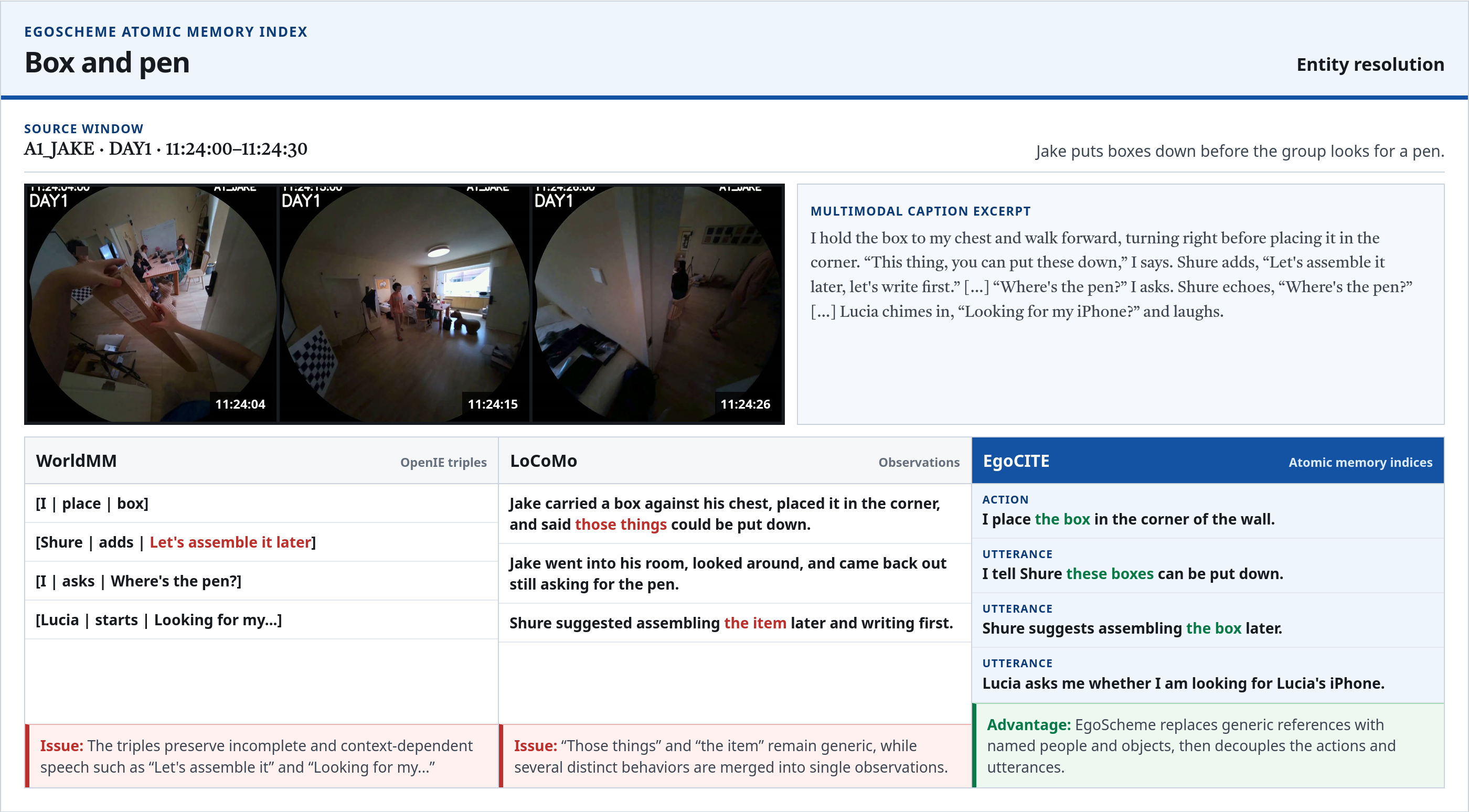}
    \caption{EgoScheme example for entity resolution. WorldMM and LoCoMo retain context-dependent references such as ``it,'' ``those things,'' and ``the item,'' whereas EgoScheme resolves the referenced boxes and decouples the corresponding actions and utterances.}
    \label{fig:egoscheme-box-and-pen}
\end{figure*}

\begin{figure*}[h!]
    \centering
    \includegraphics[width=\textwidth]{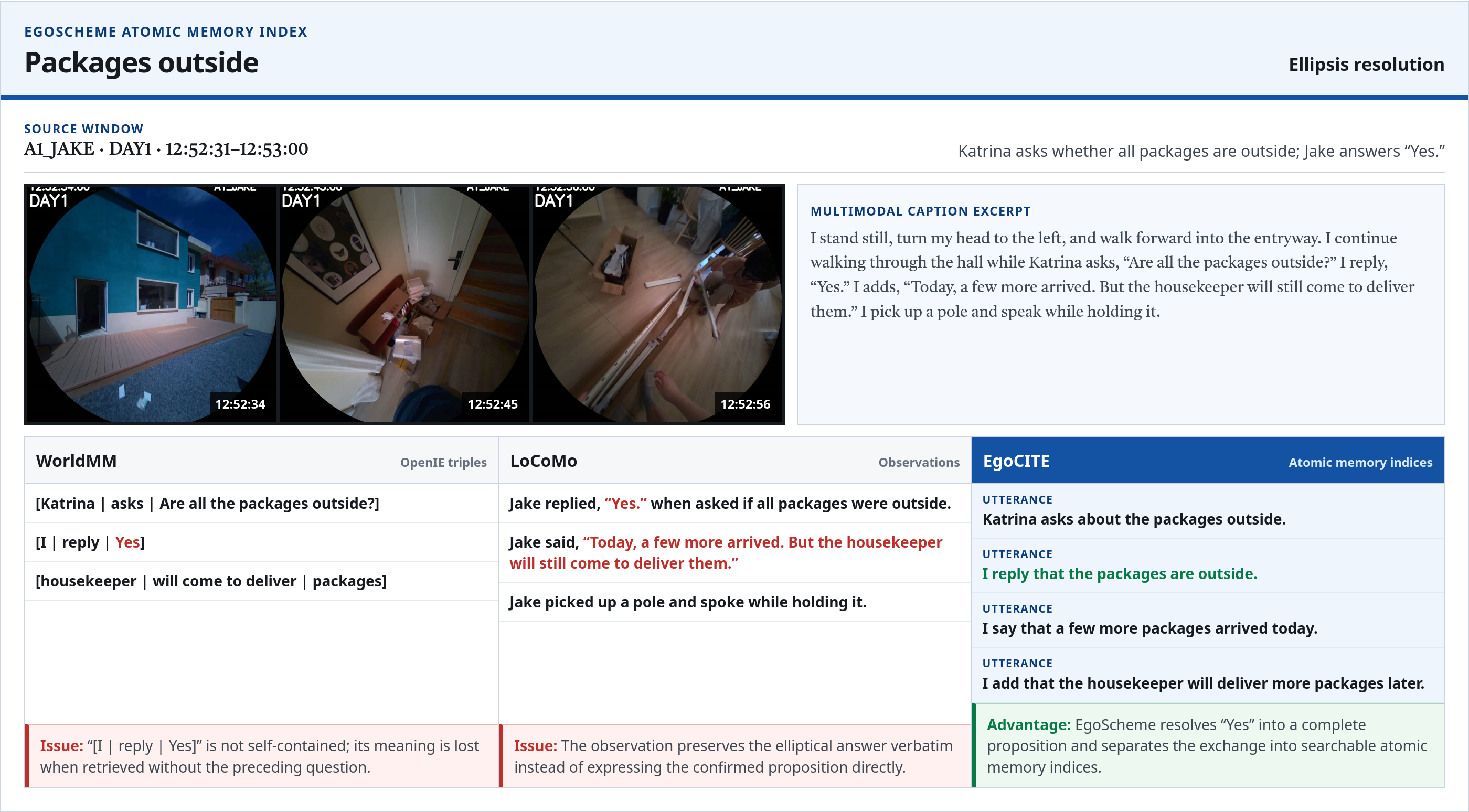}
    \caption{EgoScheme example for ellipsis resolution. WorldMM and LoCoMo preserve the elliptical reply ``Yes,'' while EgoScheme constructs the self-contained atomic memory index ``I reply that the packages are outside.''}
    \label{fig:egoscheme-packages-outside}
\end{figure*}

\begin{figure*}[h!]
    \centering
    \includegraphics[width=\textwidth]{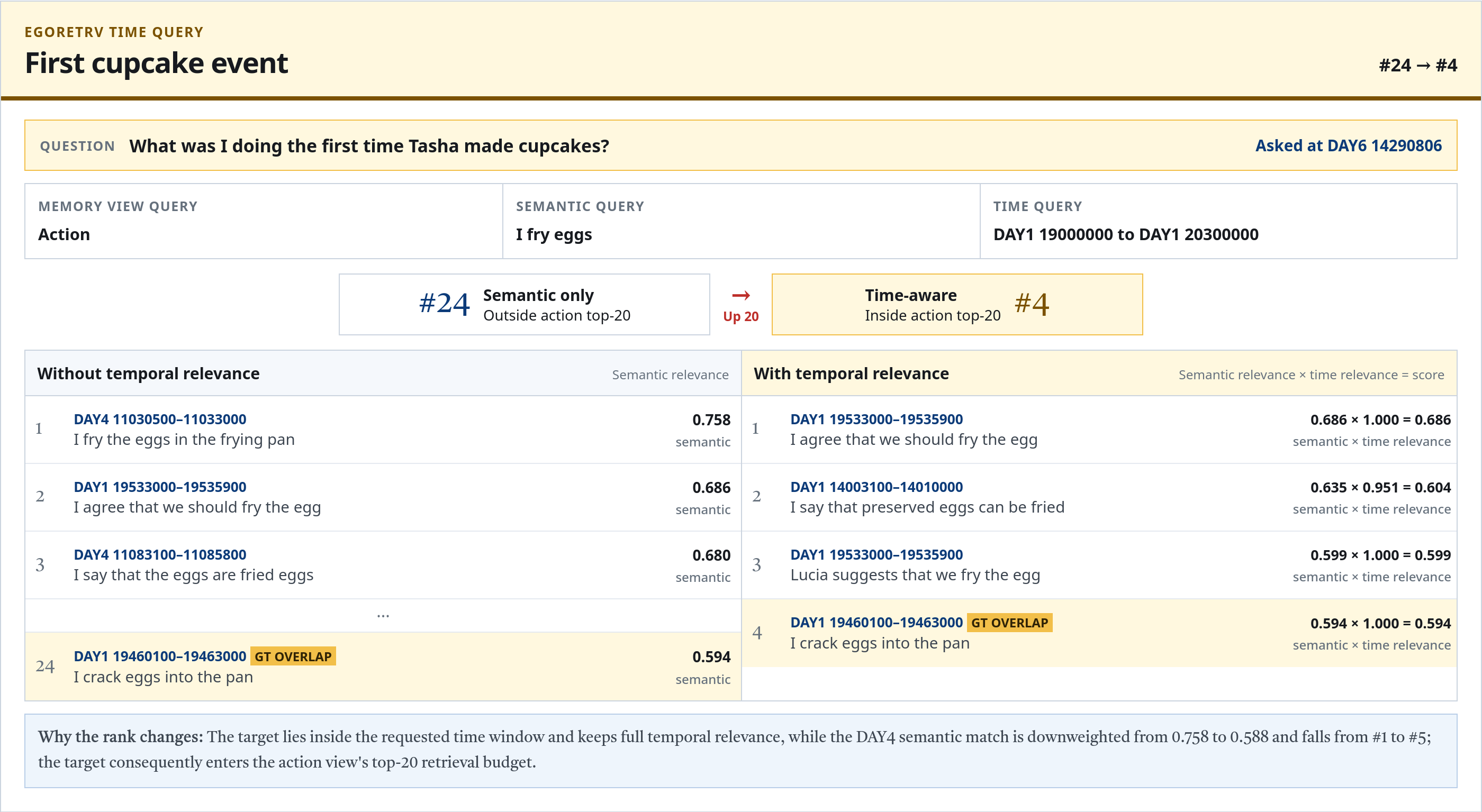}
    \caption{EgoRetrv time-query example for the first cupcake event. Temporal relevance scoring moves the target-overlapping action index from rank 24 under semantic-only retrieval to rank 4, placing it inside the action view's top-20 retrieval budget.}
    \label{fig:egoretrv-first-cupcake}
\end{figure*}

\begin{figure*}[h!]
    \centering
    \includegraphics[width=\textwidth]{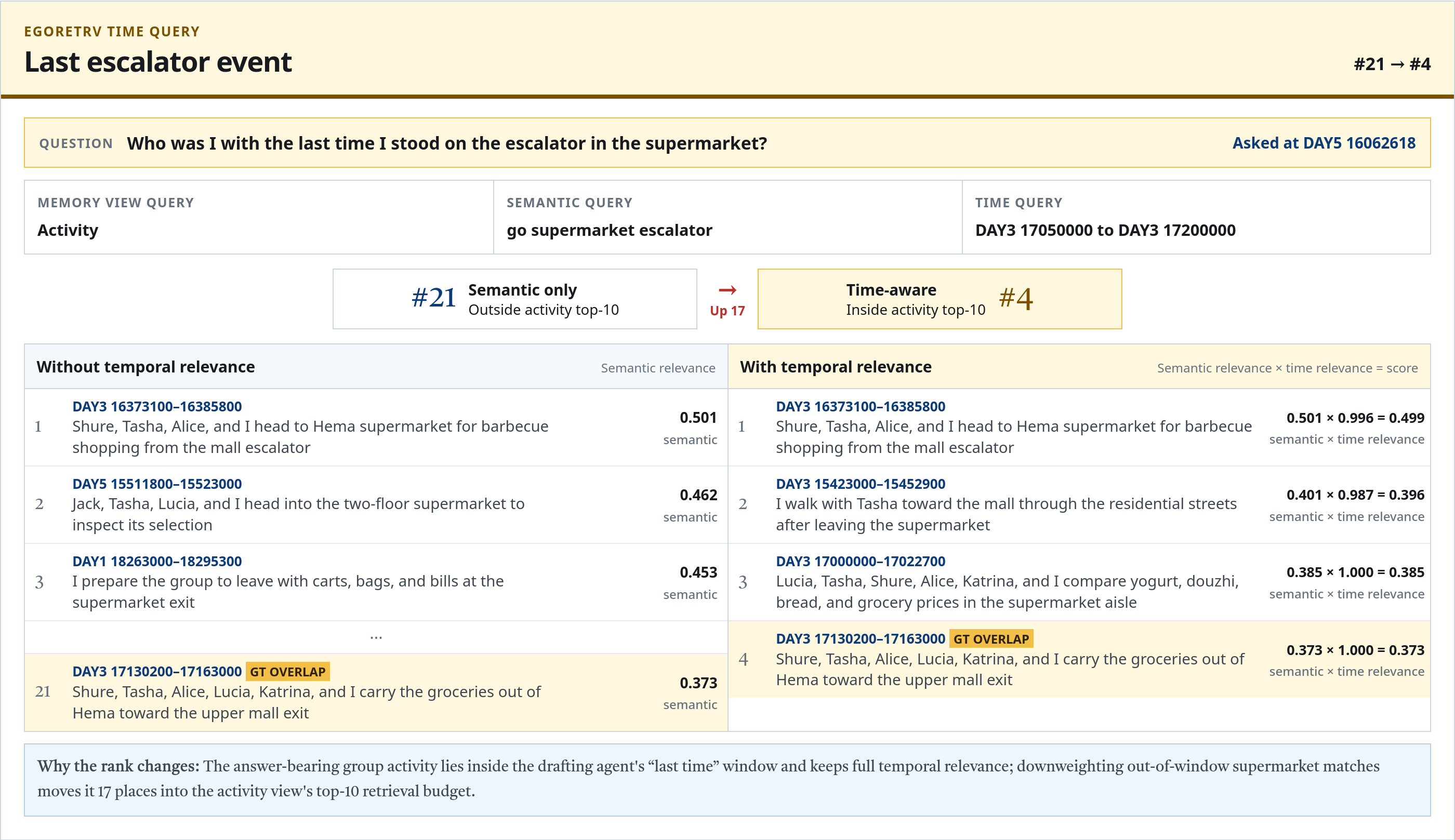}
    \caption{EgoRetrv time-query example for the last escalator event. Applying the question-conditioned time query moves the target-overlapping activity index from rank 21 to rank 4, placing it inside the activity view's top-10 retrieval budget.}
    \label{fig:egoretrv-last-escalator}
\end{figure*}

\begin{figure*}[h!]
    \centering
    \includegraphics[width=\textwidth]{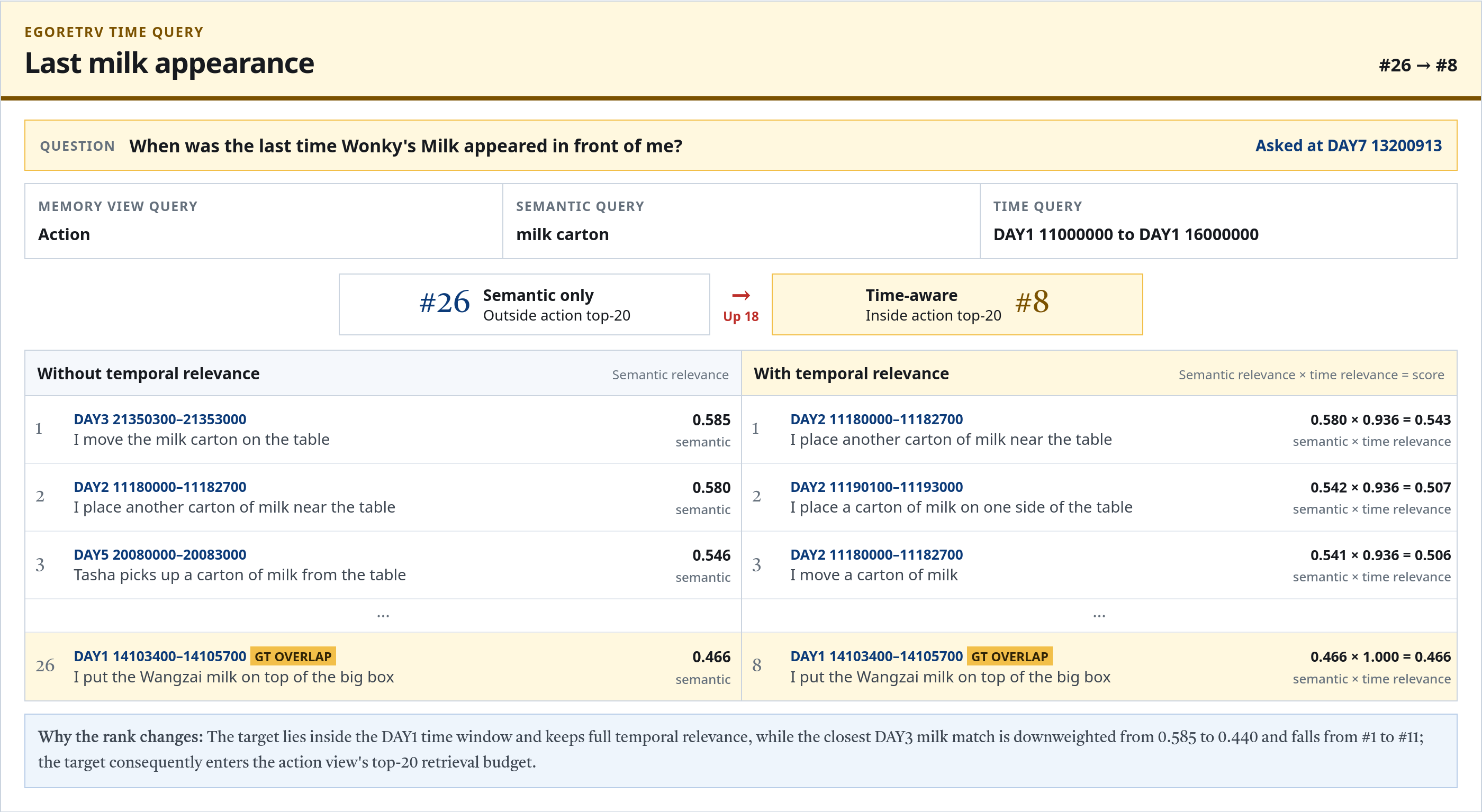}
    \caption{EgoRetrv time-query example for the last appearance of milk. Temporal relevance scoring moves the target-overlapping action index from rank 26 to rank 8, placing it inside the action view's top-20 retrieval budget.}
    \label{fig:egoretrv-last-milk}
\end{figure*}
\clearpage
\section{Prompts}

\begin{figure*}[!h]
\centering
\begin{promptbox}[Action \& Utterance Extraction Prompt.]
\begin{minted}[
    frame=none,
    framesep=3mm,
    linenos=false,
    tabsize=2,
    breaklines=true,
    breakanywhere=true,
    fontsize=\scriptsize,
    style=bw
]{text}
ONLY extract these FOUR types of entries — drop everything else:
A. HAND-OBJECT INTERACTION — someone (first-person "I" or a named housemate)
   directly handles or uses an object.
   e.g. "I fasten the shelf bracket with a screwdriver" / "Alice chops the onion".
B. BIG DISPLACEMENT — someone enters or moves into an explicitly named place.
   e.g. "I walk into the kitchen" / "Shure enters the living room" / "I sit down on the chair in living room".
C. UTTERANCE OR SPEECH — someone talks, asks, tells, suggests, agrees, or explains.
   e.g. "Jake asks Tasha about the dinner menu", "Jake agrees with Tasha on hiking this afternoon".
D. MEANINGFUL GESTURES — someone laughs, cries, etc.
   e.g. "Jake laughs out loudly".
Anything that is not type A, B, C, or D — bare gestures, posture, glances, idle — must NOT be an entry.

RULES
1. ONE entry per distinct action/fact. Never join two with "and"/"while"/"then"/"as" — split instead.
2. Each entry is 8–20 words, "subject + verb + object". Use "I" for the wearer and a real NAME for anyone else; keep named objects/places verbatim.
3. NAME every person, object, and place. Resolve all pronouns and generic words to a real NAME, "I", or the actual object/place. Drop an entry only if a referent is truly unresolvable.
4. Turn dialogue into a speech act with a SPECIFIC topic, stated as a WHAT or THAT clause. The clause MUST be complete and self-contained.
   Good: "Jake tells Tasha that dinner is ready", "I ask Alice about the shelf height"
   Bad:  "Jake tells Tasha that" (fragment), "I ask Alice about it" (not specific)
5. Drop micro-gestures and undirected motion. Keep a motion only when it reaches a named destination.

GOOD EXAMPLES:
"I message Lucia on my phone", "I fasten the shelf bracket with a screwdriver", "Shure hands me the screwdriver at desk"

BAD EXAMPLES:
"Jake says OK." -- Lack of context. Should be "Jake agrees with Tasha on hiking this afternoon".
"I use tools." -- Lack of details. Should be "I use spoon to install the shelf".
"Jake says, 'I will handle it'." -- Unclear coreference and quotes. Should be "Jake says he will handle the cleanup job".
"I'm working on computer while listening to Jake talking about weather." -- Coupled actions. Should be "I'm working on computer to edit video" and "Jake says weather will be bad tomorrow".

RESOLVE: "He picks it up" -> "Jake picks up the bottle on table"; "I walk forward" -> "I walk into the kitchen"

Output JSON only: {"actions": ["entry1", "entry2", ...]}
\end{minted}
\end{promptbox}
\caption{\texttt{EgoScheme} action and utterance resolution and extraction prompt.}
\label{fig:action-utterance-extraction}
\end{figure*}

\begin{figure*}[!h]
\centering
\begin{promptbox}[Activity Extraction Prompt]
\begin{minted}[
    frame=none,
    framesep=3mm,
    linenos=false,
    tabsize=2,
    breaklines=true,
    breakanywhere=true,
    fontsize=\scriptsize,
    style=bw
]{text}
You are given timestamped 30-second captions, indexed 0..N-1.
Segment the captions into CONTIGUOUS activity blocks, where each block represents ONE
immediate goal pursued by the wearer (and the people present).

BOUNDARY RULES:
1. Each block must span between 1 and 10 minutes of wall time
   NEVER produce a block longer than 10 minutes.
2. Blocks must be contiguous and non-overlapping; every caption index belongs to
   exactly ONE block, and indices within a block must form a consecutive run.
3. Split when the immediate goal CHANGES — a different task, different place,
   different set of people, or a clear transition.

For EACH block, write a `summary` that MUST:
- Start with "I" or a real NAME (Jake/Alice/Tasha/Lucia/Katrina/Shure);
- name the SPECIFIC object, topic, people, and place;
- state the GOAL of the block, not raw motions or sub-steps,
  e.g. "Alice, Bob, and I discuss the dinner plan", "I organize the living room for the party";
- describe ONE thing — never use "and", "while", or "then";
- be specific and rich about the expression;
- never use a vague umbrella noun (techniques, topics, things, stuff, the project, ideas).

Good: "Alice, Bob, and I plan the group dinner in the kitchen"
Good: "I edit the Earth Day promo video at my desk"
Bad:  "We discuss the techniques"   (vague umbrella noun)
Bad:  "I pick up the bottle"        (a raw sub-step, not the block's goal)
Bad:  "I organize the room"         (too simple, needs details)

Output JSON only:
{"groups": [{"indices": [0,1,2,...], "summary": "..."}, ...]}
\end{minted}
\end{promptbox}
\caption{\texttt{EgoScheme} activity extraction prompt.}
\label{fig:activity-extraction}
\end{figure*}

\begin{figure*}[!h]
\centering
\begin{promptbox}[Conversation Extraction Prompt]
\begin{minted}[
    frame=none,
    framesep=3mm,
    linenos=false,
    tabsize=2,
    breaklines=true,
    breakanywhere=true,
    fontsize=\scriptsize,
    style=bw
]{text}
You are given timestamped 30-second egocentric captions,
indexed 0..N-1. Find the CONVERSATION segments and label each with the broad
TOPIC being discussed.

Housemates — always use the real NAME: Jake, Alice, Tasha, Lucia, Katrina, Shure.

WHAT COUNTS AS CONVERSATION:
A caption is part of a conversation if the wearer ("I") is actively talking with
one or more named housemates — asking, replying, debating, joking, planning,
explaining, complaining, agreeing, etc. Reading aloud to oneself, watching a
video, or muttering does NOT count. Silent presence near someone does NOT count.

BOUNDARY RULES:
1. Each group must contain CONTIGUOUS caption indices (a consecutive run).
   Indices within a group should describe ONE coherent conversation topic.
2. A group ends when the topic shifts to a clearly different subject, OR when
   conversation pauses for >=1 minute of non-conversation activity (start a new
   group when conversation resumes).
3. Captions that are NOT part of any conversation must be LEFT OUT — do not
   force-assign them. The union of groups need NOT cover every index.
4. Empty groups list is fine if there is no conversation in this window.

For EACH group, write a `summary` that MUST:
- be 10–20 words;
- start with "I" or a real NAME;
- name every participant present in the conversation (including "I");
- state the BROAD TOPIC of the conversation, not raw motions or sub-quotes;
- describe ONE topic;
- be specific (what about the topic) — no vague umbrella nouns
  (techniques, topics, things, stuff, the project, ideas, plans).

Use plain English — write "and" normally; do NOT substitute "+", "&", "/", or "plus" for it.

Rules on connectives ("and" / "while" / "then"):
- OK to use them to list PARTICIPANTS, OBJECTS, or closely-related sub-items
  of the same topic.
- NOT OK to use them to join TWO DIFFERENT TOPICS into one summary.
  If two distinct topics were discussed, return TWO separate groups instead.

Good: "Alice, Lucia, and I plan the dinner menu and seating for tomorrow night"
Good: "Jake and I debate which model to use for the demo video"
Bad:  "We talk about stuff in the kitchen"               (vague umbrella noun)
Bad:  "I say hi to Alice"                                (gesture, not a topic)
Bad:  "Alice tells me she is going to the store"         (sub-quote, not the topic)
Bad:  "Jake + I talk about the model"                    (use "and", not "+")
Bad:  "I discuss the demo video and we plan dinner"      (two topics — split into two groups)

Output JSON only:
{"groups": [{"indices": [0,1,2,...], "summary": "..."}, ...]}
\end{minted}
\end{promptbox}
\caption{\texttt{EgoScheme} conversation extraction prompt.}
\label{fig:conversation-extraction}
\end{figure*}

\begin{figure*}[!h]
\centering
\begin{promptbox}[Drafting Agent Tool Call Schema]
\begin{minted}[
    frame=none,
    framesep=3mm,
    linenos=false,
    tabsize=2,
    breaklines=true,
    breakanywhere=true,
    fontsize=\scriptsize,
    style=bw
]{json}
{
  "tools": [
    {
      "name": "search_action_utterance",
       "description": "Search the ACTION index (atomic 30-sec observations of objects/people/physical movements/utterance). Adds matching captions to the working evidence pool with fresh integer IDs.",
      "parameters": {
        "query": {
          "type": "string",
          "required": true,
               "description": "Grounded action query (object / action+object / person+action / person+action+object / person+utterance),
          <=12 words."
         },
        "time_query": {
          "type": "string",
          "required": false,
        "description": "Optional time window in dataset format ('DAYn HHMMSSFF' or 'DAYn HHMMSSFF to DAYm HHMMSSFF'). Omit if no time signal."
        }
      }
    },
    {
      "name": "search_activity",
      "description": "Search the ACTIVITY index (5-min activity summaries). Adds matching captions to the working evidence pool.",
      "parameters": {
        "query": {
          "type": "string",
          "required": true
        },
        "time_query": {
          "type": "string",
          "required": false
        }
      }
    },
    {
      "name": "search_conversation",
      "description": "Search the CONVERSATION index (conversation-topic summaries). Use TELEGRAPHIC keyword utterance phrasing. Adds matching captions to the working evidence pool.",
      "parameters": {
        "query": {
          "type": "string",
          "required": true
        },
        "time_query": {
          "type": "string",
          "required": false
        }
      }
    },
    {
      "name": "curate_index",
      "description": "Prune the working evidence memory: keep only the listed . Use after each batch of searches to maintain a focused memory of 5-15 items. Do NOT over-curate: keeping fewer than 5 items risks discarding related evidence the QA agent needs.",
      "parameters": {
        "keep": {
          "type": "array[int]",
          "required": true,
          "description": "Evidence IDs to keep in working memory."
        }
      }
    },
    {
      "name": "answer",
      "description": "END the retrieval phase. Takes no arguments. A SEPARATE ANSWERING AGENT will be invoked with the captions in your working memory + the question + the choices, and will produce the final reasoning and answer letter itself. You do NOT pick the evidence, then call answer(). Call this ONLY after curate_index.",
      "parameters": {}
    }
  ]
}
\end{minted}
\end{promptbox}
\caption{\texttt{EgoRetrv} drafting agent tool call schema.}
\label{fig:tool-call-scheme}
\end{figure*}

\begin{figure*}[!h]
\centering
\begin{promptbox}[Sampling Agent Prompt]
\begin{minted}[
    frame=none,
    framesep=3mm,
    linenos=false,
    tabsize=2,
    breaklines=true,
    breakanywhere=true,
    fontsize=\scriptsize,
    style=bw
]{text}
You are an evidence curator for a personal egocentric memory system.
You are given a pool of candidate evidence items retrieved for a multiple-choice question.
Your job is to select the most relevant evidence items for the final QA agent.

# Selection criteria
Keep between 5 and 15 items. Do NOT over-curate: leaving only 1 or 2 items
risks discarding related evidence the QA agent needs to answer correctly.
Prefer items that:
- Directly mention the key entity / action / object / person the question asks about.
- Help resolve time qualifiers ("last", "first", "yesterday", "recently") unambiguously.
- Distinguish between the answer choices — favour items that support one choice over another.

Drop items that are:
- Redundant (same event repeated → keep the most informative one).
- Clearly off-topic or unrelated to the question and choices.

# Curation policy (apply based on question wording)
- "usually" / "often" / "typically": curate a DIVERSE set spanning the full timeline.
- "last" / "most recent" / "latest": curate the LATEST timestamps.
- "first" / "earliest": curate the EARLIEST timestamps.
- "before X" / "after X": curate items in the relevant time window relative to event X.

Call `curate_index` once with your final selection, then stop.

User message:
Question: {question}
A) {choice_A}
B) {choice_B}
...

--- CANDIDATE EVIDENCE (N items) ---
  [E1] [DAY1 HHMMSSFF-HHMMSSFF] caption text...
  [E2] [DAY1 HHMMSSFF-HHMMSSFF] caption text...
  ...

Call `curate_index` with the IDs to keep (<=15).
\end{minted}
\end{promptbox}
\caption{\texttt{EgoRetrv} sampling agent prompt.}
\label{fig:curation-agent-prompt}
\end{figure*}

\begin{figure*}[!h]
\centering
\begin{promptbox}[Response Agent Prompt]
\begin{minted}[
    frame=none,
    framesep=3mm,
    linenos=false,
    tabsize=2,
    breaklines=true,
    breakanywhere=true,
    fontsize=\scriptsize,
    style=bw
]{text}
You are an egocentric memory assistant. The person wearing the camera is asking about their own past experiences. You are given a curated set of 30-sec VIDEO captions (with timestamps) that a retrieval agent already selected as the evidence for this question.

Use ONLY these captions to answer the multiple-choice question. Do not invent details that are not in the captions.

Follow these two grounding steps before committing to a letter:
  1. GROUND every element of the question in the captions. For each key noun, person, object, or action in the question, find the caption(s) that actually describe it. If a caption does not mention it, treat it as not happening.
  2. GROUND the time qualifier (if any). Words like 'last', 'yesterday', 'recently', 'first', 'before', 'after', 'the day before yesterday' must be resolved against the caption timestamps and the memory query point — NOT against general daily schedules.

Output format (STRICT):
  - A few sentences of REASONING that cite the specific captions (by their timestamp) and explain why they support the chosen answer and rule out the others.
  - A single final line: `Answer: <letter>` where <letter> is A, B, C, or D.

User message:
[Memory query point: DAYn HHMMSSFF]

--- CURATED EVIDENCE (N raw 30-sec cap
  [DAY1 HHMMSSFF-HHMMSSFF] caption text...
  [DAY1 HHMMSSFF-HHMMSSFF] caption text...
  ...

Question: {question}
A) {choice_A}
B) {choice_B}
C) {choice_C}
D) {choice_D}

Reason from the captions above, citing
End with a single line: `Answer: <letter>`.

\end{minted}
\end{promptbox}
\caption{\texttt{EgoCITE} response agent prompt.}
\label{fig:response-agent-prompt}
\end{figure*}

\end{document}